\documentclass[journal]{IEEEtran}
\usepackage{xcolor,soul,framed} 
\usepackage{soul}
\usepackage{color, xcolor} 
\colorlet{shadecolor}{yellow}
\usepackage[pdftex]{graphicx}
\graphicspath{{../pdf/}{../jpeg/}}
\DeclareGraphicsExtensions{.pdf,.jpeg,.png}
\usepackage{array} 

\usepackage[cmex10]{amsmath}
\usepackage{mathabx}
\usepackage{placeins}
\usepackage{easyReview}
\usepackage{array}
\usepackage{mdwmath}
\usepackage{mdwtab}
\usepackage{eqparbox}
\usepackage{url}
\usepackage{multicol}
\usepackage{bbm}
\usepackage{makecell}
\usepackage{multirow}
\usepackage{inputenc}
\usepackage{cite}
\usepackage{threeparttable}
\usepackage{algorithm}  
\usepackage{algpseudocode}  
\usepackage{amsmath}  
\usepackage{endnotes}
\usepackage{booktabs}
\usepackage{float}

\begin{document}
\bstctlcite{IEEEexample:BSTcontrol}
\title{Reinforcement Learning for Scalable Train Timetable Rescheduling with Graph Representation}

\author{Peng~Yue,\IEEEmembership{}
        Yaochu~Jin,~\IEEEmembership{Fellow,~IEEE,}
        Xuewu~Dai,~\IEEEmembership{Member,~IEEE,}
        Zhenhua~Feng,~\IEEEmembership{Senior Member,~IEEE,}
        Dongliang Cui\IEEEmembership{}  

  \thanks{Peng Yue and Dongliang Cui are with the State Key Laboratory of Synthetical Automation for Process Industries, Northeastern University, Shenyang 110819, China (e-mail: pyue515@gmail.com; cuidongliang@mail.neu.edu.cn).}%
  \thanks{Yaochu Jin is with the State Key Laboratory of Synthetical Automation for Process Industries, Northeastern University, Shenyang 110819, China. He is also with the Department of Computer Science, University of Surrey, Guildford, GU2 7XH, UK (e-mail: yaochu.jin@surrey.ac.uk) \emph{(Corresponding author: Yaochu Jin)}}
    \thanks{Xuewu Dai is with Northumbria University, Newcastle upon Tyne NE1
    8ST, U.K., and also with Northeastern University, Shenyang 110819,
    China (e-mail: xuewu.dai@northumbria.ac.uk)}
  \thanks{Zhenhua Feng is with the School of Computer Science and Electronic Engineering, University of Surrey, Guildford, GU2 7XH, UK. (E-mail: z.feng@surrey.ac.uk)}}

\markboth{}
{}
\maketitle

\begin{abstract}
Train timetable rescheduling (TTR) aims to promptly restore the original operation of trains after unexpected disturbances or disruptions. Currently, this work is still done manually by train dispatchers, which is challenging to maintain performance under various problem instances. To mitigate this issue, this study proposes a reinforcement learning-based approach to TTR, which makes the following contributions compared to existing work. First, we design a simple directed graph to represent the TTR problem, enabling the automatic extraction of informative states through graph neural networks. Second, we reformulate the construction process of TTR's solution, not only decoupling the decision model from the problem size but also ensuring the generated scheme's feasibility. Third, we design a learning curriculum for our model to handle the scenarios with different levels of delay. Finally, a simple local search method is proposed to assist the learned decision model, which can significantly improve solution quality with little additional computation cost, further enhancing the practical value of our method. Extensive experimental results demonstrate the effectiveness of our method. The learned decision model can achieve better performance for various problems with varying degrees of train delay and different scales when compared to handcrafted rules and state-of-the-art solvers.
\end{abstract}

\section{Introduction}
\IEEEPARstart{H}{i}gh-speed railways provide a great convenience for passengers traveling medium and long distances~\cite{Zhou2020}. {However, many unexpected events may disturb or disrupt the normal operation of the trains, thus reducing the operation efficiency. For example, a minor malfunctioning of infrastructure tends to slightly defer the nominal arrivals and departures of multiple trains in the network, denoted as train disturbances. In extreme weather, long delays will occur and strongly affect the nominal circulation of the trains, which is defined as train disruptions~\cite{cavone2020mpc}. To deal with these delay scenarios,} human train dispatchers must modify the train schedules properly so that the trains can keep running in an efficient and ordered manner. This procedure is known as train timetable rescheduling (TTR)~\cite{Cacchiani2014, Cheng2017, Pellegrini2015}. 
Currently, although computer-aided systems, \textit{e.g.}, the centralized traffic control (CTC) system in China, can reduce the workload of dispatchers, such as train operation monitoring and command sending, it is still necessary for dispatchers to manually provide a new feasible schedule based on their experience~\cite{Dotoli2014}. 
Usually, it is extremely challenging for dispatchers to accomplish high-performance rescheduling when dealing with various delay scenarios.
In this case, the design of an effective TTR algorithm is crucial not only for improving the efficiency of railway operations but also for reducing the burden of human dispatchers.

Like most scheduling problems, the TTR problem is NP-hard~\cite{fang2015survey}. For this reason, a large body of research has been carried out to deal with this problem, which can be broadly classified into the following four categories. 
The first category is based on mathematical programming. Due to its theoretical guarantee of the solution quality, various methods have been proposed in this category~\cite{rodriguez2007constraint, Sato2013, kersbergen2016distributed, Zhang2020a}. Normally, they are pretty efficient when solving small-scale problems, but their solution time typically increases exponentially with the size of the problem, hindering their practicality to a certain degree. Many methods have been proposed to mitigate this issue, such as problem decomposition~\cite{Min2011, Zhan2021}, parallelization~\cite{Pellegrini2015}, and heuristics-assisted methods~\cite{Bettinelli2017, TornquistKrasemann2012}.
The second approach to addressing this issue is to design meta-heuristic methods, such as genetic algorithms~\cite{dundar2013train}, tabu search~\cite{kanai2011optimal} and hybrid algorithms~\cite{wang2019genetic}.  Inspired by biological intelligence, these algorithms typically exhibit a strong global search capability, enabling them to find satisfactory solutions with a shorter response time compared to the first category of methods. 

Despite the progress made in the two approaches discussed above, heuristic rules, which belong to the third category, remain prevalent in practice due to their higher computational efficiency. Popular heuristic rules include First-Come-First-Service (FCFS)~\cite{Wang2016} and First-Schedule-First-Service (FSFS)~\cite{Fang2015}. However, these heuristic rules heavily rely on domain knowledge in TTR, which leaves much room for improvement. Finally, learning-based methods, in particular reinforcement learning (RL), which are the fourth category, have been proposed for TTR in recent years. It has been proved that learning-based methods can learn advanced heuristic rules automatically, which not only significantly improves the solution quality but also maintain relatively low computational cost. Out of the above reasons, this work focuses on the RL-based approach to TTR. Notably, different from existing RL-based methods, this work contains the following main new contributions:
\begin{itemize}
\item We employ the graph representation to describe the TTR problem. In the graph, the directed edge depicts the train operation constraints of TTR. With the help of a graph neural network (GNN), an informative state can be extracted automatically, thus improving the decision-making performance.

\item  {We develop a tree search procedure to generate solutions for TTR, bringing about two main benefits. First, it facilitates the decoupling of the model's parameters from the problem's scale. Second, the track capacity constraints intrinsic to TTR can be dealt with easily by pruning infeasible branches in the constructed tree, thus ensuring the feasibility of the final scheme.}

\item {We propose a learning curriculum for model training, comprised of unbalanced instance sampling and knowledge distillation. By employing this approach, the learned model exhibits better \textit{scalability}, effectively adapting to problems with differing degrees of train delays.} 

\item We propose a simple yet efficient local search method for the TTR problem. By refining the solutions generated by the learned decision model, the quality of the solutions can be significantly improved with only a small additional computational cost, further enhancing the {scalability} of the proposed method.
\end{itemize}

The rest of this paper is organized as follows. The related work is introduced in Section~\ref{Sec_literature}. Then we describe the TTR problem and model it as a Markov decision process (MDP) in Section~\ref{MDP process}. {Section~\ref{nn and lg} elaborates our designed network architecture, while the learning algorithm and local search are discussed in Section~\ref{L_A_L_S}.} We evaluate the proposed method and report the experimental results in Section~\ref{Sec_experiments}. Last, the conclusion is drawn in Section~\ref{Sec_conclusion}.

\section{Related Work}
\label{Sec_literature}
{Inspired by the success of Alpha Go, some researchers have attempted to apply the RL approach to learn dispatching policies for TTR. One pioneering work was proposed by $\rm\check{S}$emrov \textit{et al.}~\cite{Semrov2016}, proving that the learned policies are capable of outperforming the handcrafted rules. Li \textit{et al.} provided a novel scheme based on multi-agent deep reinforcement learning~\cite{Li2022}, where the learned model can generate an optimal solution in the observed instances. Although the above methods have yielded excellent results, some challenges still remain when they are applied to real-world scenarios. One key obstacle pertains to the scalability of the approach. Specifically, we require the learned model to effectively address problems of different scales while maintaining high performance even under various levels of train delays. Notably, recent research has taken specific measures to mitigate this concern.}

{To ensure the model's scalability across various problem sizes, a typical approach involves manual extraction of key features from the railway system to create fixed-size states. For example, Khadilkar \textit{et al.} defined their state by only using track information near the train to be dispatched~\cite{Khadilkar2019}. A similar approach is also applied in \cite{Zhu2020, 9564980}.
However, these manually designed states heavily rely on domain expertise and might miss vital factors affecting TTR, such as operational conflicts between trains. Ning \textit{et al.} attempted automated state extraction using a convolutional neural network~\cite{Ning2021}, but they trained their model on a single instance, making it challenging to generalize the extracted state to other instances. To resolve the above problem, we designed a directed graph to represent the TTR problem. This approach allows for the storage of planned and actual train operation details in nodes, while the train operation conflicts are represented by directed edges in the graph. Then a GNN model is employed to extract features from the graph, which can provide more relevant information compared to manually designed features and are independent of the problem size.}

{To ensure the model's scalability to various delay scenarios, the decision model must gain prior exposure to these scenarios during the training process. Wang \textit{et al.} have made attempts in this regard, and eventually, their trained model can be applied to simple delay scenarios~\cite{9564980}, where only a single train is delayed throughout the entire timetable.
Subsequently, Peng \textit{et al.} introduced an algorithm capable of handling more complex delay scenarios, allowing for minor delays on each train~\cite{yue2023reinforcement}. However, to the best of our knowledge, there are currently no RL-based methods tailored to address train disruptions involving prolonged train delays. Our experimental results show the difficulty in training a decision model to effectively handle these diverse scenarios simultaneously. To deal with this issue, we design a learning curriculum for model training, allowing knowledge transfer from small delay scenarios to large delay scenarios, thereby enhancing the training efficiency.} 

{To enhance the model's scalability, additional search is frequently employed in the existing studies to seek solutions with improved performance. A common approach is to utilize a sampling strategy, constructing multiple solutions by selecting actions based on the probability distribution learned by the model~\cite{kool2018attention}. Bello \textit{et al.} introduced an effective search method known as active search, where a trained model is retrained to adjust the sampling probability~\cite{Bello2016}. Subsequently, Hottung \textit{et al.} proposed an improved version to reduce the computational burden by limiting the model's weight that needs retraining~\cite{hottung2021}. Despite the success of these sampling methods in maintaining scalability, they are not suited for addressing the TTR problem, particularly in the context of high-speed railways. This is due to the requirement of real-time response of TTR within seconds, making it impractical to adequately implement sampling methods. Another widely used approach in RL-based methods is local search~\cite{Zhang2020c}. In comparison to the aforementioned methods, local search offers a notable advantage in computational efficiency, making it a more suitable choice for the TTR problem. However, to the best of our knowledge, there is no local search algorithm specifically designed for the TTR problem. Hence, we propose a local search method tailored for TTR, aiming to ensure the model's scalability across various delay scenarios.}

\section{Problem Description}
\label{MDP process}

\subsection{Problem Description}
A mathematical description of the TTR problem is given in this subsection. The relevant variables are defined in Table~\ref{var_def}.

\begin{table}[!t]
\center
\caption{All notations used in the TTR problem}
\label{var_def} 
\setlength{\tabcolsep}{1mm}{
\begin{tabular}{c c p{5.6cm}}
  \toprule
 $\textbf{Type}$ & $\textbf{Symbol}$   & $\textbf{Definition}$ \\ 
  \midrule
  \multirow{7}{*}{\begin{tabular}[c] {@{}c@{}} Decision\\variables \end{tabular}} & $a_{k,i}$    & Arrival time of train $k$ at station $i$ \\  
   & $d_{k,i}$    & Departure time of train $k$ at station $i$ \\  
   & $e_{k,i}$    & Delay penalty of train $k$ at station $i$ \\  
   & $y_{k_1,k_2,i}$  & If train $k_1$ departs earlier than train $k_2$ at station $i$, $y_{k_1,k_2,i}=1$, otherwise, $y_{k_1,k_2,i}=0$. \\ 
   & $z_{k,i,p}$  & If train $ k $ occupies track $p$ at station $i$, 
   $z_{k,i,p}=1$, otherwise, $z_{k,i,p}=0$. \\  \midrule
   \multirow{15}{*}{Parameters} & $I$                   & Number of stations \\
   & $K$                   & Number of trains   \\
   & ${a}_{k,i}^{*}$       & Original arrival time of train $k$ at station $i$ \\
   & ${d}_{k,i}^{*}$       & Original departure time of train $k$ at station $i$ \\
   & $rd_{i}$              & Minimum dwell time of trains at station $ i $\\
   & $h$                   & Minimum headway between trains \\
   & $P_{i}$  & Track capacity at station $ i $ \\
   & $rt_{k,i}$  & Minimum running time of train $k$ between station $i$ and $i+1$ (\textit{i.e.}, section $ i $) \\
   & $M$             & A sufficiently large positive value \\
   & $\lambda$       & Penalty factor for the early arrival of trains\\
   & $e^*_{k, i}$    & Train delays that have occurred for train $k$ at station $i$ \\  
  \bottomrule
\end{tabular}}
   \begin{tablenotes}[r]
     \item[1] We set the value of $M$ to be 10 times greater than the maximum value observed in the planned train schedule when using the Gurobi solver to solve the TTR problem.
   \end{tablenotes}
\end{table}

The main purpose of TTR is to recover normal train operation as soon as possible, which minimized the following objective function:
\begin{equation}
\label{objective}
J=\sum_{i=1}^{I}\sum_{k=1}^{K}\left(e_{k,i}\right) \\ 
\end{equation}
where $e_{k, i}$ denotes the delay penalty of train $k$ at station $i$, and it is defined as:
\begin{equation}
\label{objective2}
e_{k,i}=\left\{\begin{array}{ll}
a_{k,i}-a_{k,i}^* & \text{if $a_{k,i} > a_{k,i}^*$} \\
\lambda*(a_{k,i}^*-a_{k,i}) & \text{else}
 
\end{array}\right.    
\end{equation}
where $a_{k, i}$ is the arrival time of train $k$ at station $i$, $a^*_{k, i}$ is the planned one of train $k$ at station $i$, {$\lambda$ represents the relative weighting of early arrivals relative to late arrivals of trains}\footnote{{$\lambda$ is set to 0.3 in the experimental section, consistent with the parameter setting in \cite{zhu2017bi}.}}. It is noteworthy that this defined non-linear objective function can be linearized as:
\begin{equation}
\label{e_def}
\left\{\begin{array}{l}
e_{k,i}\geq a_{k,i}-a_{k,i}^* \\
e_{k,i}\geq\lambda*(a_{k,i}^*-a_{k,i})\\
\end{array}\right.
\end{equation}

Due to the limitation of physical conditions, train operations should satisfy the following constraints.
\subsubsection{The minimum operation time constraint}
When train $k$ passes through one station or one section (between two stations), a certain operation time will surely be spent:
 \begin{equation}  \label{eq5}
 \begin{cases}
     a_{k,i+1}-d_{k,i}\geq{rt}_{k,i}, & {\forall i=1,\ldots,I-1; k=1,\ldots,K;} \\ 
     d_{k,i}-a_{k,i}\geq {rd}_i, & {\forall i=1,\ldots,I; k=1,\ldots,K;}
  \end{cases} 
  \end{equation} 

where $d_{k,i}$ is the departure time of train $k$ at station $i$, ${rt}_{k,i}$ is minimum running time of train $k$ in section $i$, ${rd}_{i}$ is minimum dwell time of trains at station $i$.
\subsubsection{The planned and delay constraints}
The trains should not depart earlier than their planned time to ensure the normal passenger flow, and due to unexpected events, some trains have to be delayed, which can be formulated as:
\begin{equation}\label{planned_cons}
\begin{cases}
    d_{k, i}-{d}_{k, i}^{*} \geq 0, \\
    a_{k, i}-{a}_{k, i}^{*} \geq e^*_{k,i}, \\
\end{cases}  {\forall i=1,\ldots,I; k=1,\ldots,K;}
\end{equation}
where ${d}_{k, i}^{*}$ is the original departure time of train $k$ at station $i$, and ${e}_{k, i}^{*}$ is the train delay of train $k$ at station $i$.
\subsubsection{The headway constraints}
For these trains running in the same section, overtaking is not allowed due to the limited tracks, while a time interval $h$ should be maintained for any two trains, as described by the following constraints:
\begin{align}
\begin{cases}
a_{k_2, i+1}-a_{k_1, i+1} \geq h-M(1-y_{k_1,k_2,i}), \\
a_{k_1, i+1}-a_{k_2, i+1} \geq h-M*y_{k_1,k_2,i}, \\
d_{k_2, i}-d_{k_1, i} \geq h-M(1-y_{k_1,k_2,i}), \\
d_{k_1, i}-d_{k_2, i} \geq h-M*y_{k_1,k_2,i},
\end{cases} \notag \\
\hfill  {\forall i=1,\ldots,I; k=1,\ldots,K;} \label{eq6}
\end{align}
where $y_{k_1,k_2,i}=1$ means that train $k_1$ will depart station $i$ earlier than train $k_2$.

\subsubsection{The track occupation constraints}
For any two trains (\textit{e.g.,} train $k_1$ and train $k_2$), when they intend to occupy the same track $p$ at station $i$, a minimum time interval $h$ is required: 
\begin{align}
&a_{k_2,i}-d_{k_1,i} \geq h-M*(3-y_{k_1,k_2,i-1}-z_{k_1,i,p}-z_{k_2,i,p}), \notag \\
&\qquad \qquad \qquad \qquad \qquad \quad   {\forall i=2,\ldots,I; k=1,\ldots,K;} \label{track_cap}
\end{align}
where $z_{k_1,i,p}=1$ ($z_{k_2,i,p}=1$) denotes train $k_1$ ($k_2$) will occupy the track $p$ at station $i$.
\subsubsection{Natural constraints}
Based on the variable definition, the defined variables should satisfy the following:
\begin{align}
&\ y_{k_1,k_2,i}\in \{0,1\},\   {\forall i=1,\ldots,I; k_1, k_2=1,\ldots,K;} \notag \\
&\begin{cases}
    a_{k, i} \geq 0, \\
    d_{k, i} \geq 0,
\end{cases} \quad \ \  {\forall i=1,\ldots,I; k=1,\ldots,K;} \notag \\
&\begin{cases}
  z_{k,i,p}\in \{0,1\},  \\
  \sum_{ {p =1, \ldots, P_i}} z_{k,i,p} = 1.
\end{cases} \notag \\
& \qquad \quad \quad  {\forall p=1,\ldots,P_i; i=2,\ldots,I; k=1,\ldots,K;} \label{na_cons}
\end{align}
\begin{figure*}[!t]
	\begin{center}
		\includegraphics[width=0.6\textwidth]{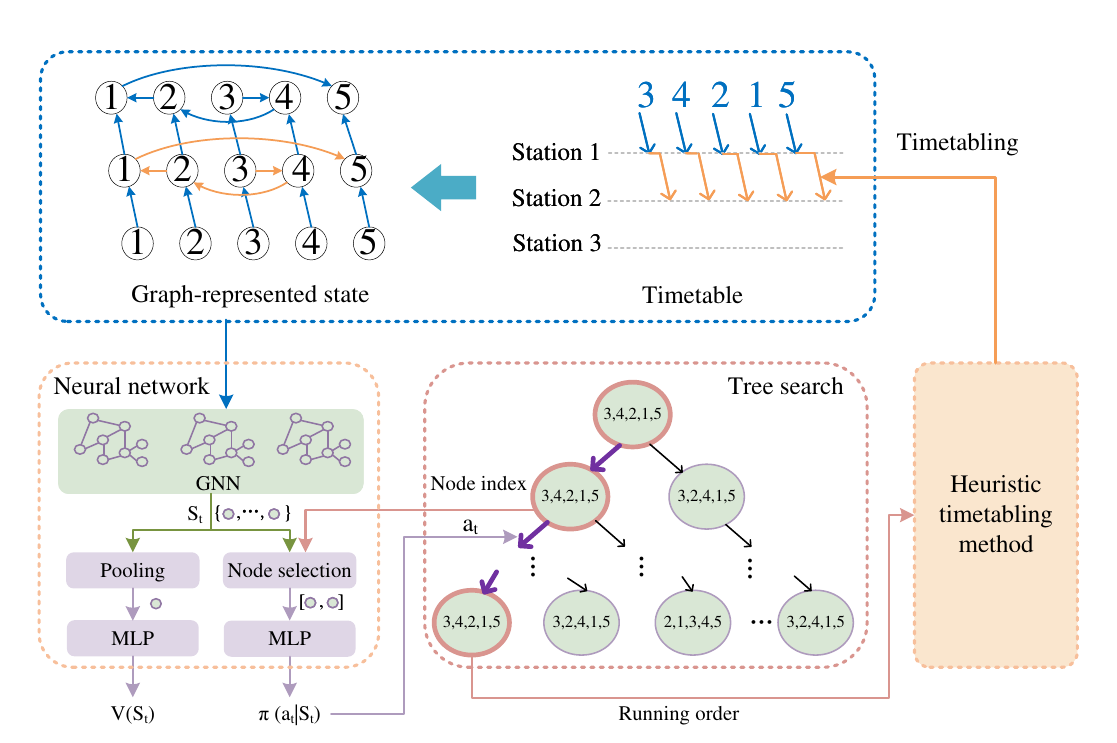}\\
		\caption{ {The designed MDP for constructing the TTR's solution.}}\label{MDP}
		\label{fig1}
	\end{center}
\end{figure*}

\section{ {Markov Decision Process for TTR}}
 {When applying the RL algorithm to solve the TTR problem, the first step is to build a suitable Markov decision process (MDP). In this study, we followed the approach outlined in \cite{Wang2021, yue2023reinforcement}, where the decision model only needs to determine the running order between the trains, and the specific time is obtained by using existing heuristics. Figure~\ref{MDP} provides a detailed illustration of our designed decision process.}

 {Specifically, a TTR's solution is constructed section by section. For each section, a timetable is first transformed into a directed graph. The GNN module then extracts a state $S_t$ from this graph. Subsequently, the tree search process is implemented, guided by the neural network, to determine the running order of trains. At each decision branch, action $a_t$ represents the routing selection. Finally, an existing heuristic method generates the operation scheme of the trains based on the determined running order. This iterative process continues until the whole timetable is constructed. Note that for clarity, we only show the detailed information needed at each stage in this figure, omitting the state transition and reward calculation during the tree search, which will be elaborated in Subsection~\ref{st_rd}}.

 {One thing we need to mention is that this designed decision process ensures the model's scalability in terms of the problem size. Firstly, the GNN module is inherently independent of the number of nodes in the graph (corresponding to the problem size). Then, during the tree search process, the fixed-length node embedding is obtained through node selection and fed into the Multi-Layer Perceptron (MLP) layer. This way, the parameters of the neural network are totally independent of the problem size.}

 {Regarding the heuristic timetabling method, it is also employed in the past studies~\cite{Wang2021, yue2023reinforcement}, but we make some revisions to allow for early arrival of the trains. The details can be found in Algorithm \ref{alg: heuristic}.
}
\begin{algorithm}[!b]
\caption{A heuristic timetabling method}  
\label{alg: heuristic}  
\begin{algorithmic}[1]  
    \Require  
    \Statex $O_a^i$: The given order of trains in section $i$
    \Statex $\{a_{k, i}\ \text{for}\ k=1,\dots, K\}$: The arrival time of trains at station $i$
    \Ensure  
    \Statex $\{a_{k, i}, a_{k, i+1}\ \text{for}\ k=1,\dots, K\}$: The updated arrival time 
    \Statex $\{d_{k, i}\ \text{for}\ k=1,\dots, K\}$ : The departure time of trains at station $i$
    \State Sort the trains based on $O_a^i$ to obtain $\Omega_s$
    \For {$k$ \textbf{in} $\Omega_s$} 
        \State Calculate the number of trains at station $i$ when
        \Statex \quad \quad  train $k$ arrives at time $ a_{k, i}$, noted as $n_{k}$
        \If {$n_{k} = P_i $}
            \State Delay the arrive time $ a_{k, i}$ of train $k$ by:
            \Statex \qquad \qquad  $ a_{k, i} = h + d_{g_1, i} $ where there are $(P_i-1)$ trains 
            \Statex \qquad \qquad  between train $ k $ and train $ g_1 $.
            \State  {Allocate the track occupied by train $g_1$ to train $k$.}
        \Else
            \State  {Allocate the unoccupied track randomly to train $k$.}
        \EndIf
        \State Calculate the time $ d_{k, i}$ and $ a_{k, i+1} $ by Eq. \eqref{det_dep} and \eqref{det_arri}.
    \EndFor
    \For{$k$ \textbf{in} $\Omega_s$}
        \State Calculate buffer time $B_h(g_2,k,i+1)$ by Eq.~\eqref{b1}
        \If{$B_h(g_2,k,i\raisebox{0mm}{+}1)<0$}
            \State  {Get train set $\Omega_{p}$ that can advance their arrival time.} 
            \State   {Advance arrival time of the train in $\Omega_{p}$ by Eq.~\eqref{b4}.}
            \If{$(\Delta_{g_2})<|B_h(g_2,k,i\raisebox{0mm}{+}1)|$}
                \State  {Delay arrival time $a_{k_f, i+1}$ by Eq.~\eqref{b5}}
            \EndIf    
        \EndIf
    \EndFor
\end{algorithmic}  
\end{algorithm}

In Algorithm~\ref{alg: heuristic}, the train operation scheme for each train is determined one by one according to the given running order $O_a^i$. Specifically,  {for train $k$, we first check whether there are available tracks for it to stop in station $i$. If not, its arrival time must be delayed until a train (corresponding to train $g_1$) leaves the station. It also meant that train $k$ and $g_1$ have to occupy the same track on station $i$, referring to lines 3-9 in Algorithm~\ref{alg: heuristic}. For its departure time $d_{k, i}$, we employ equation \eqref{det_dep} to obtain a feasible and earliest time, hoping to reduce its delay as much as possible. 
\begin{equation}
\label{det_dep}
  d_{k, i} = \max \left(d_{k, i}^{*}, a_{k,i}+rd_i, d_{g_2, i}+ h\right) 
\end{equation}
where train $ g_2 $ is the preceding train of train $k$. The same manner can also be used to obtain the arrival time at the next station~\cite{yue2023reinforcement}, but it requires that the train can not arrive at stations earlier than its planned time, which is not necessary in practice. To fill this gap, we provide a new manner to obtain a more flexible solution. Specifically, we first calculate the earliest time for each train to arrive at the next station, ignoring the operating conflicts between trains, referring to Eq.~\eqref{det_arri}. 
\begin{equation}
\label{det_arri}
  a_{k, i+1} = \max (a_{k, i}^{*}, d_{k, i}+rt_{k,i}) 
\end{equation}
To validate whether the obtained time is feasible, we will calculate the buffer time $B_h(g_2,k,i\raisebox{0mm}{+}1)$ between trains\footnote{ {Here, we only need to use the buffer time to check if the minimum interval is satisfied between trains (i.e., headway constraints), since the constraints involving a single train have already been considered in Eq.~\eqref{det_arri}.}}.
\begin{equation}\label{b1}
B_h(g_2, k,i\raisebox{0mm}{+}1) = a_{k, i\raisebox{0mm}{+}1}-a_{g_2, i}-h
\end{equation}
where train $g_2$ is the preceding train of train $k$. 
If there is a conflict between adjacent trains (i.e., $B_h(g_2, k, i+1)<0$ ), minor revisions will be implemented to dissipate conflicts, referring to lines 15-19.}

 {There are two main steps to complete the revisions. The first step is to advance the arrival time of the trains in $\Omega_{p}$, denoting a set of trains before a conflict occurs\footnote{The number of the trains in $\Omega_{p}$ is determined by $\lambda$. For example, when $\lambda$ is 0.3, then the number of trains that can be advanced is 3, as it incurs a smaller cost in objective $J$ to delay one train than to advance four trains.}. For this reason, we need to calculate the adjustable time for these trains iteratively while satisfying their own constraints, calculated by Eq.~\eqref{b2}.
\begin{equation}
\label{b2}
h_{aj}(k_p) \raisebox{0mm}{=}
\min\big(B_r(k_p,i), h_{aj}(k_p\raisebox{0mm}{-}1)\raisebox{0mm}{+}B_h(k_p\raisebox{0mm}{-}1, k_p,i\raisebox{0mm}{+}1)\big) 
\end{equation}
where $B_r(k_p,i)=a_{k_p, i+1}-a_{k_p, i}-rt_{k,i}$ represents the supplement time of train $ k_p$ in segment $i$. For the earliest arrival train in $\Omega_{p}$, $h_{aj}(k_p)$ is set as $B_r(k_p,i)$.
Next, we advance the arrival time of these trains in $\Omega_{p}$ by $\Delta_{k_p}$:
\begin{equation}\label{b4}
a_{k_p, i+1} = a_{k_p, i+1}-\Delta_{k_p}
\end{equation} 
Specifically, for train $g_2$, i.e., the preceding train involved in conflicts, its actual adjustment $\Delta_{g_2}$ is the minimum value between the adjustable time $h_{aj}(g_2)$ and the violation $(-B_h(g_2,k,i+1))$:
\begin{equation} \label{b3_1}
    \Delta_{g_2}=\min\left({-B_h(g_2,k,i\raisebox{0mm}{+}1), h_{aj}(g_2)}\right)
\end{equation}
To avoid triggering new conflicts, we also need to adjust other preceding trains in $\Omega_p$ iteratively: 
\begin{equation}
    \Delta_{k_p\raisebox{0mm}{-}1}=\max\big(\Delta_{k_p}\raisebox{0mm}{-}B_h(k_p\raisebox{0mm}{-}1, k_p, i\raisebox{0mm}{+}1), 0\big)
\end{equation}
}
 {If the above adjustment is not enough to resolve the conflict (i.e., $\Delta_{g_2}<|B_h(g_2,k,i\raisebox{0mm}{+}1|$), the second step will be implemented to delay the arrival time of the trains in $\Omega_s(k\raisebox{0mm}{:}\text{end})$ iteratively, denoting the set of trains after the conflict occurs. The specific updating process is shown in Eq.~\eqref{b5}.
\begin{equation}\label{b5}
a_{k_f, i+1} = \max\big(a_{k_f, i+1}, (a_{k_f\raisebox{0mm}{-}1, i+1}+ h)\big)
\end{equation} 
where train $k_f\in\Omega_s(k\raisebox{0mm}{:}\text{end})$ indicates the train operating after the conflict between $g_2$ and $k$.}

\section{ {Network Architecture}}
\label{nn and lg}
 {In this section, we will initially present the designed network architecture for TTR. Subsequently, we will provide a detailed description of its application in constructing solutions for TTR, encompassing the GNN-based state and tree search. Additionally, we will elaborate on the state transitions and how rewards are calculated during the tree search process.}

\begin{figure}[!t] 
\begin{center}
    \includegraphics[width=0.5\textwidth]{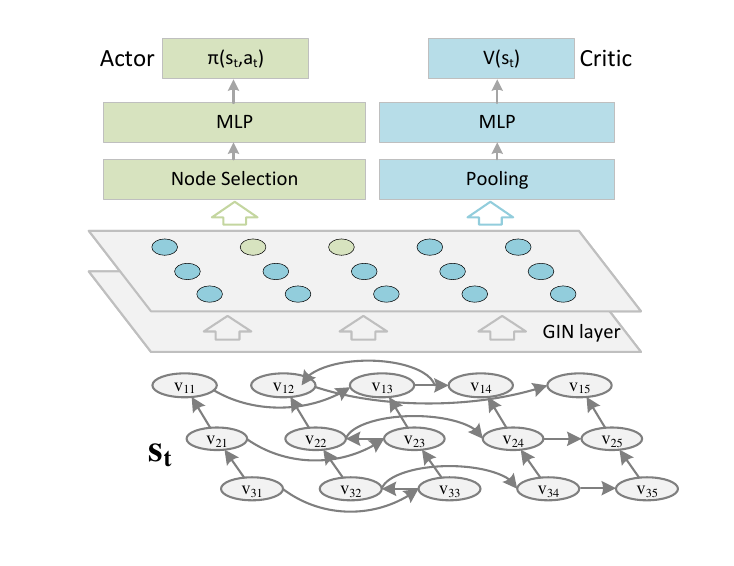}\\
    \caption{The network architecture for the decision model.}
    \label{fig: network_arc}
\end{center}
\end{figure}

 {The whole network architecture is shown in Fig.~\ref{fig: network_arc}, which can be divided into two parts: the GIN layer and the task-specific layer. Firstly, the GIN layer is applied to extract the state features. Subsequently, the critic subnetwork undertakes the state evaluation, denoted as $V(S_t)$, while the actor subnetwork is engaged in action selection $\pi(a_t|s_t)$. These components will be thoroughly elucidated in the subsequent subsection.}

\subsection{GNN-Based State Representation}
 {In this subsection, we will describe how to describe a TTR instance as a directed graph, thus extracting the state information by using the GIN layer.}
\subsubsection{Graph Representation for TTR}
\begin{figure}[!t]    
\begin{center}
    \includegraphics[trim = 8mm 0mm 5mm 8mm, clip,width=0.5\textwidth]{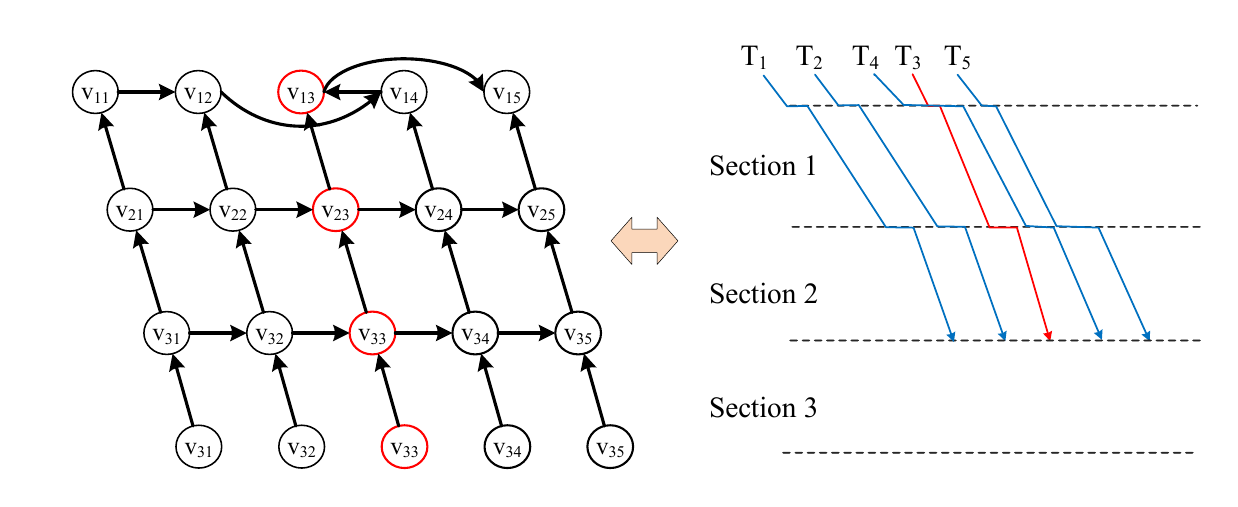}\\
    \caption{An example of modeling the TTR as a directed graph.}\label{fig:directed graph}
\end{center}
\end{figure}

In the TTR problem, it is challenging for the dispatcher to determine an effective running order between trains. The main reason is that there are so many factors influencing the scheduling results, \textit{e.g.}, the original schedule, the actual schedule, the type of locomotive, and the coupling between trains, so it is also non-trivial for a handcrafted state to grasp this information completely. Therefore, we attempt to describe this information by using a directed graph, and a GNN can be used to grasp key information automatically. Figure~\ref{fig:directed graph} shows an example of a designed graph. 

The graph can be defined as $G = (V, E)$, which consists of two components: nodes $V=\{v_{k,i}:[\delta_{k, i}, a_{k, i}^*, L_{k, i}]\}$ and edges $E$. Node $v_{k, i}$ represents the arrival event of train $k$ at station $i$ and records three features, including train delay $\delta_{k, i}$ (calculated by Eq.~\eqref{upda_e}), planned arrival time $a_{k, i}^*$ and a binary indicator $L_{k, i}$ indicating whether it has been rescheduled ($L_{k, i}=1$ means it has been rescheduled). 
\begin{equation}\label{upda_e}
\delta_{k,i} =
\begin{cases}
  {\max} \big(0,\delta_{k,i-1}-B_r(k,i\raisebox{0mm}{-}1)\big) & \text{if $L_{k,i} = 0$} \\
 \text{Calculated by Algorithm}~\ref{alg: heuristic} & \text{if $L_{k,i} = 1$},
 \end{cases}
\end{equation}
where $B_r(k, i\raisebox{0mm}{-}1)$ indicates the supplement time of train $k$ in section $(i\raisebox{0mm}{-}1)$. It can be found that the train delay is not always accurate when $L_{k, i}$ is 0,  but it is efficient in improving the training efficiency.

 {The edge connecting two nodes describes the constraint relationship between adjacent events, which is similar to the existing AG graph. But there exist significant differences between them. In the AG graph, the edge direction is consistent with the sequence of events, proceeding from the antecedent event to the subsequent one. Conversely, in our designed graph, when two adjacent events describe the same train at different stations, we reverse the edge's direction, making it point from the subsequent event to the antecedent one (\textit{e.g.,} the edge between node $v_{13}$ and $v_{23}$). This modification enables the decision model to obtain future information from subsequent nodes, thus better determining the running order of trains.}

\subsubsection{State Extraction}
Based on the graph, a GNN model can be used to extract the state feature automatically. In the GNN family, we adopt the Graph Isomorphism Network (GIN)\cite{xu2018powerful} to achieve this due to its strong discriminative power. 

Specifically, given a graph $G = (V, E)$, the node representation can be extracted in an iterative way. In each iteration, the node embedding is updated by aggregating the information from its neighborhoods, which is formulated as:
\begin{equation}\label{gnn}
h_v^{(k)}\raisebox{0mm}{=}\text{ReLU}\Bigg(\text{BN}\bigg(\text{MLP}_{\theta_k}^{(k)}\Big((1\raisebox{0mm}{+}\epsilon^{(k)})h_v^{(k\raisebox{0mm}{-}1)}
    \raisebox{0mm}{+}\sum_{u \in \mathcal{N}(v)}h_u^{(k\raisebox{0mm}{-}1)}\Big)\bigg)\Bigg),\\
\end{equation}
where $h_v^{(k\raisebox{0mm}{-}1)}$ indicates the node embedding of node $v$ in the iteration $(k\raisebox{0mm}{-}1)$, and the raw feature $h_v^{0}$ is defined as $h_v^{0} = [\delta_{k, i}, a^*_{k, i}]$. $\mathcal{N}(v)$ is the neighborhoods of the nodes $v$, $\text{MLP}_{\theta_k}^{(k)}(\cdot)$ is a multi-layer perceptron (MLP) with parameter $\theta_k$, $\epsilon^{(k)}$ is learnable scalar, $\text{BN}(\cdot)$ stands for batch normalization, $\text{ReLU}(\cdot)$ is the rectified linear unit. 

\subsection{ {Running order determination by using tree search process}}
\begin{figure}[!t]   
\begin{center}
    \includegraphics[trim = 8mm 0mm 5mm 8mm, clip, width=0.4\textwidth]{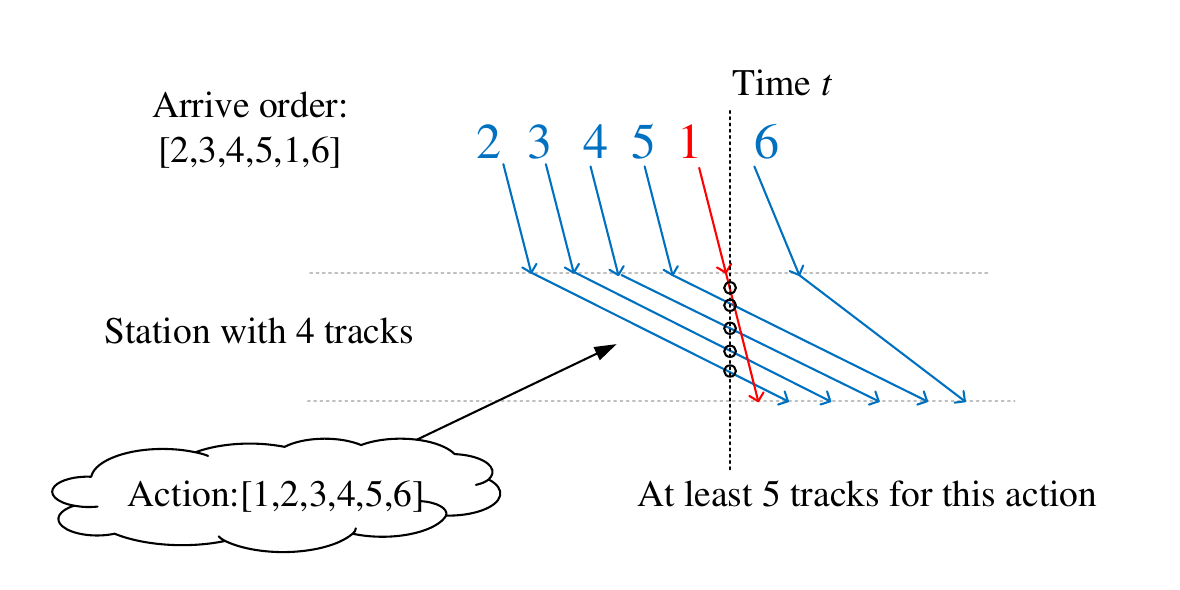}\\
    \caption{An example of an infeasible order.}\label{fig:example_unfeasible}
\end{center}
\end{figure}

 {Given the extracted node embedding, we need to provide a feasible running order for trains.} A prevalent method for sequence determination in prior studies~\cite{Wang2021, Zhang2020b} involves incrementally incorporating one item (i.e., trains) into the partial sequence, but this process ignores the inherent constraints specific to TTR, denoted as track capacity constraints. Figure~\ref{fig:example_unfeasible} shows a simple example of an infeasible running order.

\begin{figure}[!b]    
\begin{center}
    \includegraphics[width=0.5\textwidth]{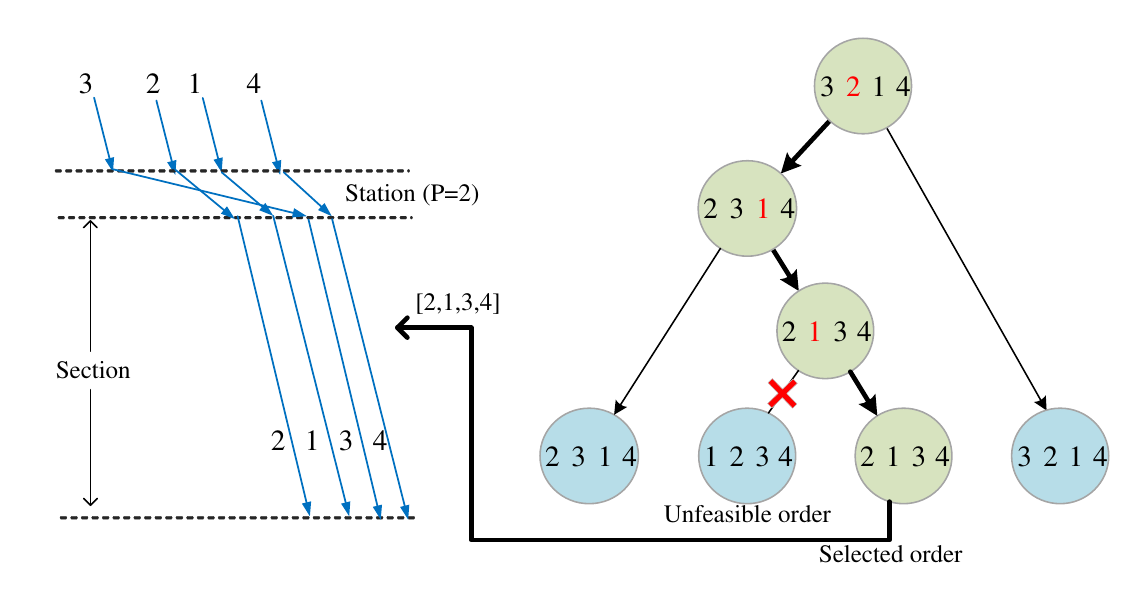}\\
    \caption{ {The tree search process for generating a feasible running order of trains.}}
    \label{fig:decision_tree}
\end{center}
\end{figure}

It can be found that each train is not allowed to overtake more than ($P_i-1$) trains at station $i$, otherwise, the number of trains will exceed the number reserved. To address this issue, this study designs a tree search process to obtain feasible solutions for TTR, as shown in Fig.~\ref{fig:decision_tree}.

The running order between trains in a section can be determined by getting a route in the constructed tree, and the task of the decision model is thus transformed into giving priority between two adjacent trains at each bifurcation point. There are mainly two benefits. First, the tracking capacity constraints can be addressed directly by pruning the infeasible branches. Second,  {the fixed-length node embedding is obtained by concatenating the two nodes related to the trains to be dispatched, thus decoupling the model's parameters from the problem size.} Since this procedure is similar to the dispatcher's behavior, we employ the dispatcher's knowledge to construct the tree, thereby enhancing the search efficiency. The specific process is illustrated in Algorithm~\ref{build_tree}.

\begin{algorithm}[!t]  
\caption{Construction of the search tree}  
\label{build_tree}  
\begin{algorithmic}[1]  
    \Require  
    \Statex $O_a^{i\raisebox{0mm}{-}1}$: The actual running order of trains in section $i\raisebox{0mm}{-}1$;
    \Statex $O_p^i$: The planned running order of trains in section $i$ 
    \Ensure  
    \Statex The constructed tree;
    \State $O_0 = O_a^{i\raisebox{0mm}{-}1}$\qquad\qquad \% The root node of the tree
    \State \textbf{\% Determine the overtaken trains $\Omega_{ot}$}
    \State $O_{ac}=\text{copy}(O_{a}^{i\raisebox{0mm}{-}1})$
    \For{$k$ in $O_p^i$}
        \State Find the location of train $k$ in $O_{ac}$:
        \Statex \qquad \quad $loc_k = \text{find}(O_{ac}, k)$
        \If{$loc \neq 1$}
            \State Add $k$ into set $\Omega_{ot}$
        \EndIf
        \State Delete $k$ from $O_{ac}$
    \EndFor
    \State \textbf{\% Create the child nodes iteratively}
    \State $O_f=O_0$  \qquad \% $O_f$ is the father node
    \For{$k$ in $O_a^{i\raisebox{0mm}{-}1}$}
        \If{$k$ in $\Omega_{ot}$}
            \State Find the proceeding train of train $k$ in $O_f$:
            \Statex \qquad \qquad \ $g_2=O_f(\text{find}(O_f,k)-1)$ 
            \While{$g_2$ is not empty}
                \If {$\text{find}(O_{p}^i,g_2)<\text{find}(O_{p}^i,k$)}
                    \State \textbf{Break}
                \Else
                    \State Create two child nodes $O_{c1}$ and $O_{c2}$:
                    \Statex  \qquad \qquad \qquad \quad $O_{c1}=O_f$
                    \Statex  \qquad \qquad \qquad \quad  $O_{c2}=\text{Swap}(O_f, k,g_2)$
                    \State $O_f = O_{c2}$
                    \State  $g_2=O_f(\text{find}(O_f,k)-1)$ 
                \EndIf
            \EndWhile
        \EndIf
    \EndFor
\end{algorithmic}
\end{algorithm}

First, we define the running order of trains in the last section (\textit{i.e.}, $O_a^{i\raisebox{0mm}{-}1}$) as the root node $O_0$ in the tree. Second, the overtaken trains $\Omega_{ot}$ are identified iteratively by comparing the positions of the same trains in $O_a^{i\raisebox{0mm}{-}1}$ and $O_p^{i}$, referring to lines 4-10.
Third, we create the child nodes in the tree iteratively by swapping the position between overtaken train $k$ and its preceding train $g_2$ (\textit{i.e.,} $\text{Swap}(O_f, k, g_2)$). It should be noted that train $k$ cannot overtake the train with a higher priority in planned order $O_p^i$, referring to lines 17-18. When all overtaken trains in $\Omega_{ot}$ are adjusted, the whole tree is constructed. It can be found that the constructed tree will help the overtaken trains to restore their original positions as soon as possible. The motivation stems from the actions of train dispatchers. In practice, train dispatchers have a preference for selecting the departure order closer to the original one, which is due to the fact that it has less impact on the subsequent scheduling tasks, such as the rolling stock scheduling~\cite{wang2018passenger}. Furthermore, it also mitigates the effects on passengers who require transfers.

 {Recall that the decision model's task is to select a route at each decision branch. Thus, we design the upper layer of the neural network as follows:}

For the actor subnetwork, the concatenation operator $[\cdot,\cdot]$ is firstly implemented in the node selection layer for two nodes $v_{i, k_1}$ and $v_{i, k_2}$ to be compared, and  {an MLP layer and sigmoid function $\sigma(\cdot)$ are then applied to determine their priorities:}
\begin{equation}\label{policy}
     \pi_{\psi}(a_{t}|s_t) = \sigma(\text{MLP}_{\psi}([h_{v_{i, k_1}}^{K}, h_{v_{i, k_2}}^{K}])). 
\end{equation}
 {where $h_{v_{i, k_1}}^{K}$ and $h_{v_{i, k_2}}^{K}$ are the selected node embedding.}

For the critic subnetwork, a global representation $h_g$ of the entire graph is obtained by using the average pooling layer. We apply an MLP to obtain the state value $V(s_t)$:
\begin{equation}\label{pooling}
\left\{\begin{array}{l}
     h_g = \text{Pool}(\{h_v^{K}:v \in V\})=\sum_{v\in V}h_v^K /|V| \\
     V(s_t)_{\phi}=\text{MLP}_{\phi}(h_g).
    \end{array}\right.
\end{equation}

\subsection{State Transition and Reward Definition}
\label{st_rd}
\subsubsection{State Transition}
At each decision step, the decision model will provide a priority between two adjacent trains. If the given order is changed, the constraint relationships are also updated, as reflected by the change of edge in the graph. Once the decision model has completed a route search of the built tree in a section, an entire running order of the trains can be obtained. At this point, it is necessary to update the node features to provide a more precise prediction of train delays (\textit{i.e.,} $\delta_{k, i}$), using Algorithm~\ref{alg: heuristic}.
\begin{figure}[!b]    
\begin{center}
    \includegraphics[width=0.45\textwidth]{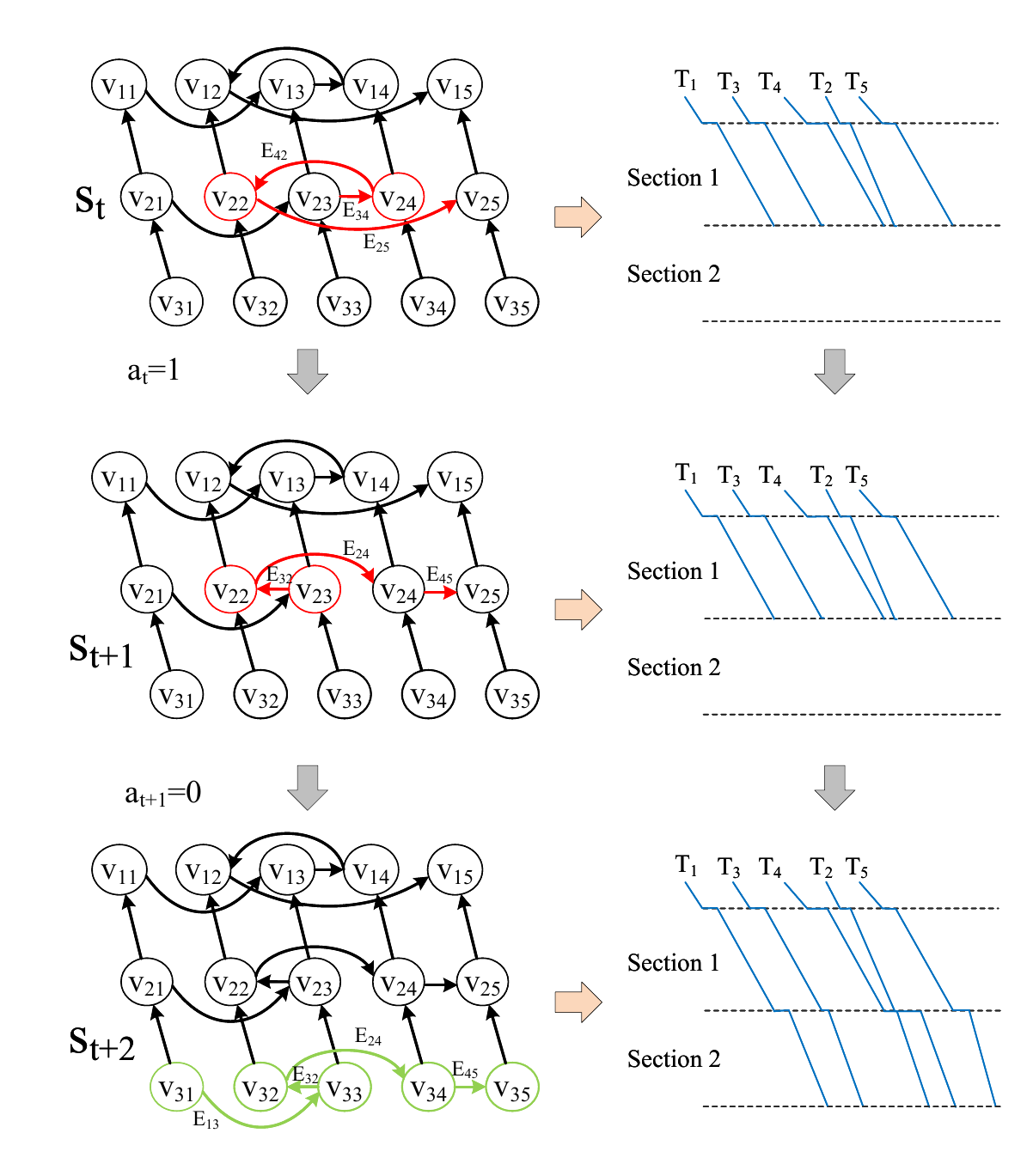}\\
    \caption{An example of state transition.}
    \label{state_tran1}
\end{center}
\end{figure}

For easier understanding, Fig.~\ref{state_tran1} depicts the process of state transitions. At the decision step $t$, the decision model provides the priority between trains $T_2$ and $T_4$, and the final decision is to reverse their order (\textit{i.e.}, $a_t=1$), so related constraints have to be changed. For example, the red edges $E_{25}$ (from node $v_{22}$ to $v_{25}$), $E_{42}$ and $E_{34}$ in the graph are deleted, and some new edges are added, including $E_{32}$, $E_{24}$ and $E_{45}$. In the next step, the decision model chooses to maintain the priority between trains $T_2$ and $T_3$ (\textit{i.e.}, $a_{t+1}=0$), so the corresponding edge remains unchanged. 
Since the entire tree search procedure has been completed, Algorithm~\ref{alg: heuristic} can be applied to update the train schedule in this section. Accordingly, some new edges (\textit{i.e.}, green edges) reflecting the running order will be added to the graph, and the node features also need to be updated. To be more specific, $L_{w, i}$ for the green nodes is assigned a value of 1, and $\delta_{k, i}$ is calculated using Eq.~\eqref{upda_e}.

\subsubsection{Reward Definition}
In this study, we define an unbiased reward for objective $J$, which is calculated by Eq.~\eqref{reward}.
\begin{equation}\label{reward}  
    r_t= \begin{cases}
         (\hat{J}^{t-1}-\hat{J}^{t})/(K*I) & \text{if node feature is updated} \\
          \ 0 & \text{else},
         \end{cases}
\end{equation}
where $\hat{J}^{t-1}$ and $\hat{J}^{t}$ are the predicted objective value in the current state and the updated state, respectively, and $K*I$, as the description of problem size, is used to reduce the variance of reward.
When the entire running order is determined by the decision model, the reward is defined as the decrease in the objective function $J$. Otherwise, the reward is assigned a value of 0.

\section{Learning Algorithm and Local Search}
\label{L_A_L_S}
\subsection{ {Learning Algorithm}}
 {In this study, we aim to train a generic model scalable to various delay scenarios, including train disturbances and disruptions. However, our experiments reveal that current RL algorithms fall short in training such a scalable decision model. To address this, we design a learning curriculum (LC) to improve the training performance. The detailed implementation can be found in Algorithm~\ref{training}.}

\begin{algorithm}[!b]
	\caption{ {Proposed Learning algorithm}}
	\label{training}  
	\begin{algorithmic}[1]  
    \Require $\{\theta, \psi, \phi\}$: The initial parameter of the decision model
    \Ensure
    Trained decision model $\{\theta^*, \psi^*, \phi^*\}$
    \For{$stage \ sta = 1, \cdots, 2$}  
    \For{$episode \ epi = 1, \cdots, \text{Max\_epi}$}
        \If{$sta == 1$}
            \State Generate a problem instance randomly
        \Else
            \State Generate a instance based on distribution $P_{LC}$
        \EndIf
        \While{True}
        \State Transform the TTR instance as a graph $G$
        \State State extraction by using GIN layers 
        \State Get a running order $a_t$ by using the tree search
        \State Generate a partial solution using Algorithm~\ref{alg: heuristic}
        \If{the whole timetable is generated}
            \State \textbf{break}
        \EndIf
        \EndWhile
        \For{$Epoch \ epo = 1, \cdots, \text{Max\_epo} $}
            \State Calculate loss item $L(\theta, \psi, \phi)$ using Eq~\eqref{para_update}.
            \If{$sta == 2$}
                \State Calculate loss item $L_4(\theta, \phi)$ using Eq~\eqref{il_eq}.
                \State $L(\theta, \psi, \phi)$ = $L(\theta, \psi, \phi) + w_4 \cdot L_4(\theta, \phi)$
            \EndIf
            \State Update $\{\theta, \psi, \phi\}$ by using Adam optimizer
        \EndFor
    \EndFor
    \EndFor
	\end{algorithmic}  
\end{algorithm}

 {We divide the training process into two stages. In the first stage, we employ the vanilla PPO algorithm to train the decision model on small delay scenarios. Then, in the second stage, the learned model undergoes retraining using our proposed learning curriculum to address the large delay scenarios. Our learning curriculum improves upon the vanilla PPO method in two key aspects: instance sampling and loss definition. The rationale behind this is to transfer knowledge obtained from handling small delay scenarios (a relatively easy task) to effectively handling large delay scenarios (a more challenging task). }

 {Regarding instance sampling, the PPO algorithm generates TTR instances randomly, assigning each train a uniform random delay. In contrast, we introduce an unbalanced sampling weight $P_{LC}=[p_s, p_l] (p_s>p_l)$, where $p_s$ is the sampling probability for small delay scenarios (train disturbances), and $p_l$ corresponds to large delay scenarios (train disruptions), as outlined in lines 3-7 of Algorithm~\ref{training}. Concerning the loss definition, we introduce an additional loss term $L_4(\theta,\phi)$ to implement knowledge distillation, referring to lines 19-22 of Algorithm~\ref{training}. Specifically, the original loss function $L(\theta, \psi, \phi)$ used in PPO algorithm is defined as:}
\begin{equation} \label{para_update}
\begin{cases}
L_1({\theta,\psi}) = -\sum_t^T{\min{\left(\mathcal{E}_t\hat{A}_{t},clip(\mathcal{E}_t,1-\epsilon,1+\epsilon)\hat{A}_{t}\right)}}\\
L_2({\theta,\phi}) = \frac{1}{T}\sum_t^T(\sum_t^T \gamma^t r_t - V(s_t))^2\\
L_3(\theta,\psi) = -\sum_{t}^T H([\pi_{\theta,\psi}(s_t), 1-\pi_{\theta,\psi}(s_t)]) \\
L({\theta,\psi,\phi}) = w_1 L_1({\theta,\psi}) + w_2 L_2({\theta,\phi})  + w_3 L_3({\theta,\psi}),
\end{cases} 
\end{equation}
where $\mathcal{E}_t=\frac{\pi(a_{t}|s_t)}{\pi_{old}{(a_{t}|s_t)}}$ is the probability ratio between the old policy $\pi_{old}$ and current policy $\pi$, and $\hat{A}_t = \sum_t^T \gamma^t r(s_t,a_t)-V(s_{t})$ is an estimator of the advantage function at timestep $t$, and $H(\cdot)$ is the entropy loss function. In this function, $L_1({\theta,\psi})$ is responsible for improving the policy steadily, and $L_2({\theta,\phi})$ is used to estimate the state value $V(s_t)$, and $L_3({\theta,\psi})$ is employed to ensure the exploratory power of the model. 

 {The extra loss $L_4({\theta,\psi})$ is defined as follows:}
\begin{equation}
\label{il_eq}
L_4({\theta,\psi}) = H(\pi_{S}(a_t|s_t), \pi(a_t|s_t)) 
\end{equation}
 {where $\pi_{S}(a_t|s_t)$ represents the learned policy from the first stage, and $\pi(a_t|s_t)$ represents the current policy.}

\subsection{ {Local Search for TTR}}
 {Using the learning curriculum and tree search, the learned decision model can address any TTR instances directly logically, involving various degrees of train delays and different problem sizes. However, maintaining peak performance across all instances remains challenging. To enhance the model's performance, we introduce a local search method during the application stage. This method is akin to the commonly used 2-opt heuristics~\cite{costa20a} in TSP, as detailed in Algorithm~\ref{local_search}.}

\begin{algorithm}[!b] 
	\caption{ {Local search}}  
	\label{local_search}  
	\begin{algorithmic}[1]
		\Require  
		\Statex The search times Max\_{ls}
		\Statex A rescheduled timetable.
		\Ensure  
		\Statex A updated rescheduled timetable
            \For{$ls=1,\cdots,\text{Max\_{ls}}$}
                \While{True}
                    \State Randomly select a overtaken train $k$ and station $i$.
                    \If{$y_{k,adj(k),i} \neq y_{k,adj(k),i-1}$}
                        \State \textbf{Break}
                    \EndIf
                \EndWhile
                \State Order swap between train $k$ and $adj(k)$ at station $i$.
                \State Implement Algorithm \ref{alg: heuristic} locally to rapidly estimate 
                \Statex \ \  \ \ \   the objective change, denoted as $\Delta J$.
                \If{$\Delta J<0$}
                    \For{$i_2=i,\ldots, I$}
                        \State Order swap between train $k$ and $adj(k)$ at              \Statex \qquad \qquad \ \  station $i_2$.
                        \State Estimate the new objective change $\Delta J$.
                        \If{$\Delta J>0$}
                            \State \textbf{Break}
                        \EndIf
                    \EndFor
                \EndIf
            \EndFor
	\end{algorithmic}  
\end{algorithm}

\begin{figure}[!t]    
\begin{center}
    \includegraphics[width=0.5\textwidth]{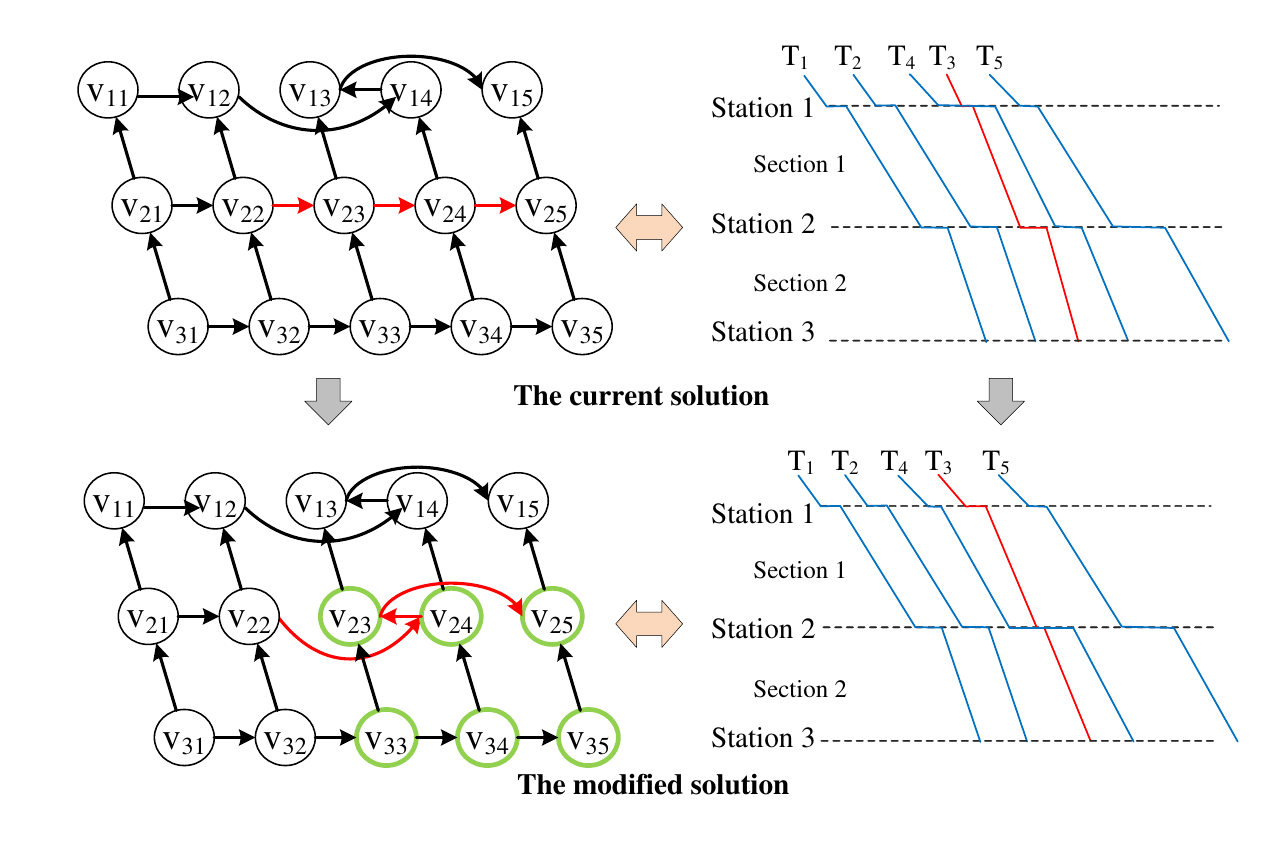}\\
    \caption{An example of local search.}\label{example_ls}
\end{center}
\end{figure}

 {The local search is mainly divided into three steps. First, we randomly select an arrival event of the train from the entire timetable, denoted as $v_{i, k}$. If train $k$ swaps its positions with its adjacent train $adj(k)$ compared to the previous station $i-1$ (i.e., $y_{k, adj(k), i} \neq y_{k, adj(k), i-1}$), the second step will be implemented to reorder train $k$ and $adj(k)$, making it consistent with the order at station $i-1$. Subsequently, Algorithm~\ref{alg: heuristic} is employed again to revise the specific arrival/departure time due to the reordering operation. Notably, this revision procedure only starts after the adjustment event and stops when the conflict is resolved, thus saving the computational cost. The change in objective function $J$ is then calculated and denoted as $\Delta J$, corresponding to lines 8-9 in Algorithm \ref{local_search}. The final step is to do further exploration if the performance of the scheme is improved, i.e., $\Delta J<0$. To be more specific, the same reordering operation will be performed again between train $k$ and $adj(k)$  at subsequent stations until the solution quality cannot be further improved, see lines 10-19 of the Algorithm. To help the reader understand its procedure, we also give a simple example in Fig.~\ref{example_ls}.}

 {In this example, node $v_{23}$ is firstly selected by executing the first step, since its position has changed compared to the first station. Next, we proceed to the second step to rearrange its order with its adjacent train $v_{24}$, depicted by the red edge in the graph. By using Algorithm~\ref{alg: heuristic}, the arrival time can be revised, denoted by the green nodes of the graph. If the revised scheme exhibits improved performance, we proceed to exchange the order between nodes $v_{33}$ and $v_{34}$ in the third step. If not, one local search iteration is completed.}
\section{Experimental Results}
\label{Sec_experiments}
\subsection{Datasets and Experimental Settings}
To make it practical for real-time decision-making, the decision model must be pre-trained on diverse problem instances, enabling it to address unexpected disturbances rapidly. However, due to the privacy concerns associated with railway operation data, obtaining an adequate and representative set of delay scenarios is challenging. In this study, we generate augmented instances from limited instances by using Algorithm~\ref{augmented_data}.
\begin{algorithm}[!b] 
	\caption{Data augmentation for TTR instances}  
	\label{augmented_data}  
	\begin{algorithmic}[1]
		\Require  
		\Statex An actual train schedule $TS_a$ for TTR; 
		\Statex Disturbance parameters $[\tau_1,\tau_2,\tau_3]$
		\Ensure  
		\Statex The problem instances $TS_b$
		\State $TS_b = TS_a+M_{r}*\tau_1$
        \State $M_{eq}=J_{I} \cdot \text{diag}(\text{range}(max(TS_b)))$
		\State $TS_b = TS_b + M_{eq}$
		\State Calculate the running order $O_b$ based on $TS_b$
		\State  $TS_b$=Algorithm \ref{alg: heuristic}($TS_b, O_b$)
		\State Assign an initial delay $ad\in[0,\tau_2]$ to each train, and set the minimum operation time $r_{k,t} \in [\tau_3 \cdot r_t^*,r_t^*]$ 
	\end{algorithmic}  
\end{algorithm}

Specifically, we first construct a raw instance $TS_b$ by introducing random noise into train timetable $TS_a$. Then we design a matrix $M_{eq}=J_I\cdot\text{diag}(\text{range}(max(TS_b)))$, where $J_I$ is all-one matrix, and $\text{range}(max(TS_b))$ stands for the arithmetic sequence with an upper bound as $max(TS_b)$. By adding this matrix into $TS_b$, the trains will be uniformly distributed over schedule $TS_b$, which is a common feature of the train schedule. To obtain a feasible train schedule, Algorithm \ref{alg: heuristic} is adopted to eliminate the conflicts in $TS_b$. Regarding the train delay, we assign a random delay $ad\in[0,\tau_2]$ to each coming train, and a minimum running time $r_{k,t}\in [\tau_3 \cdot r_t^*,r_t^*]$ is also set randomly to reflect the differences in train types, where $r_t^*$ is the original parameters in $TS_a$.

\begin{figure}[!t]
	\begin{center}
		\includegraphics[width=0.49\textwidth]{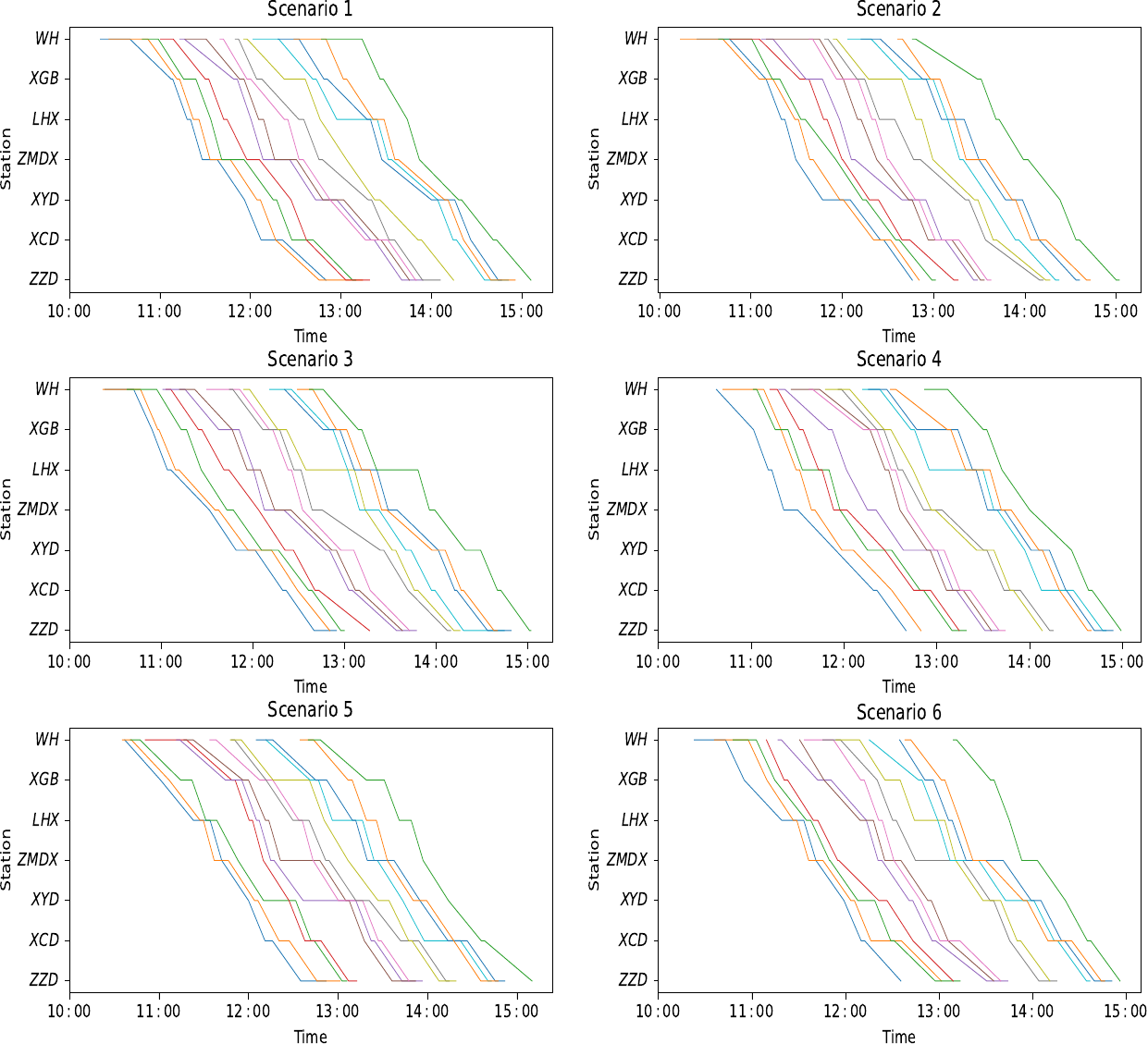}\\
		\caption{The augmented TTR instances.}
		\label{train_data}
	\end{center}
\end{figure}

According to the real scenarios, the disturbance parameter $\tau_1$ is set as 20 (min), which can ensure the diversity of the original schedule as much as possible, and some of the generated examples are shown in Fig.~\ref{train_data}. $\tau_2$ is set to 60 and 180 for train disturbances and train disruptions, respectively. Parameter $\tau_3$ is set to 0.3.

The hyperparameters of the proposed algorithm are set as presented in Table~\ref{table: hyperpara}. Notably, we assign parameter $w_3$ different values for different scenarios. Given that the values of $L_1$ and $L_2$ increase with the severity of train delays, a larger value of $w_3$ is required to balance the exploration in large-delay scenarios. We use a computer to conduct all experiments with a specification of an Intel Core i9-10940X CPU and 32GB RAM. The method is implemented in Python with PyTorch.

\begin{table}[!t]
\center
\caption{The hyperparameters used in the proposed method}
\renewcommand\arraystretch{1.2}
\label{table: hyperpara}
\setlength{\tabcolsep}{0.7mm}{
    \begin{tabular}{c l c}
        \toprule
        Type & Hyperparameter & Value\\ 
        \midrule
        \multirow{6}{*}{Network} & Number of GIN layers & 1 \\
        & Hidden dimensions of GIN layers & 128 \\
        & Number of MLP layers for Critic & 2 \\
        & Hidden dimension of MLP for Critic & 128\\ 
        & Number of MLP layers for Actor & 2  \\
        & Hidden dimension of MLP for Actor & 128 \\ 
        \midrule
        \multirow{7}{*}{Training} & Discount rate $\gamma$ & 0.9 \\ 
        & Clipping rate $\epsilon$ & 0.2 \\
        & The weights $w_1,w_2,w_3,w_4$ & [2, 2, (0.1, 0.03), 100]\\
        & Number of training episodes &  5e4\\
        & Learning rate & 1e-4 \\
        & Optimizer & Adam  \\
        & Number of updates for one episode & 10 \\
        & Unbalanced sampling weight & [0.8, 0.2] \\
        \midrule
        \multirow{1}{*}{Application} & Number of local search & 50 \\
        \bottomrule
\end{tabular}}
\end{table}

 {Recall that the model training process is divided into two stages. In the first stage, the vanilla PPO algorithm is employed to train the decision model under train disturbances. Subsequently, in the second stage, the learned model undergoes retraining using our proposed learning curriculum to handle train disruptions. Therefore, in the subsequent section, we will verify the algorithm's effectiveness step by step in this order.}
\begin{figure}[!tb]    
\begin{center}
    \includegraphics[width=0.48\textwidth]{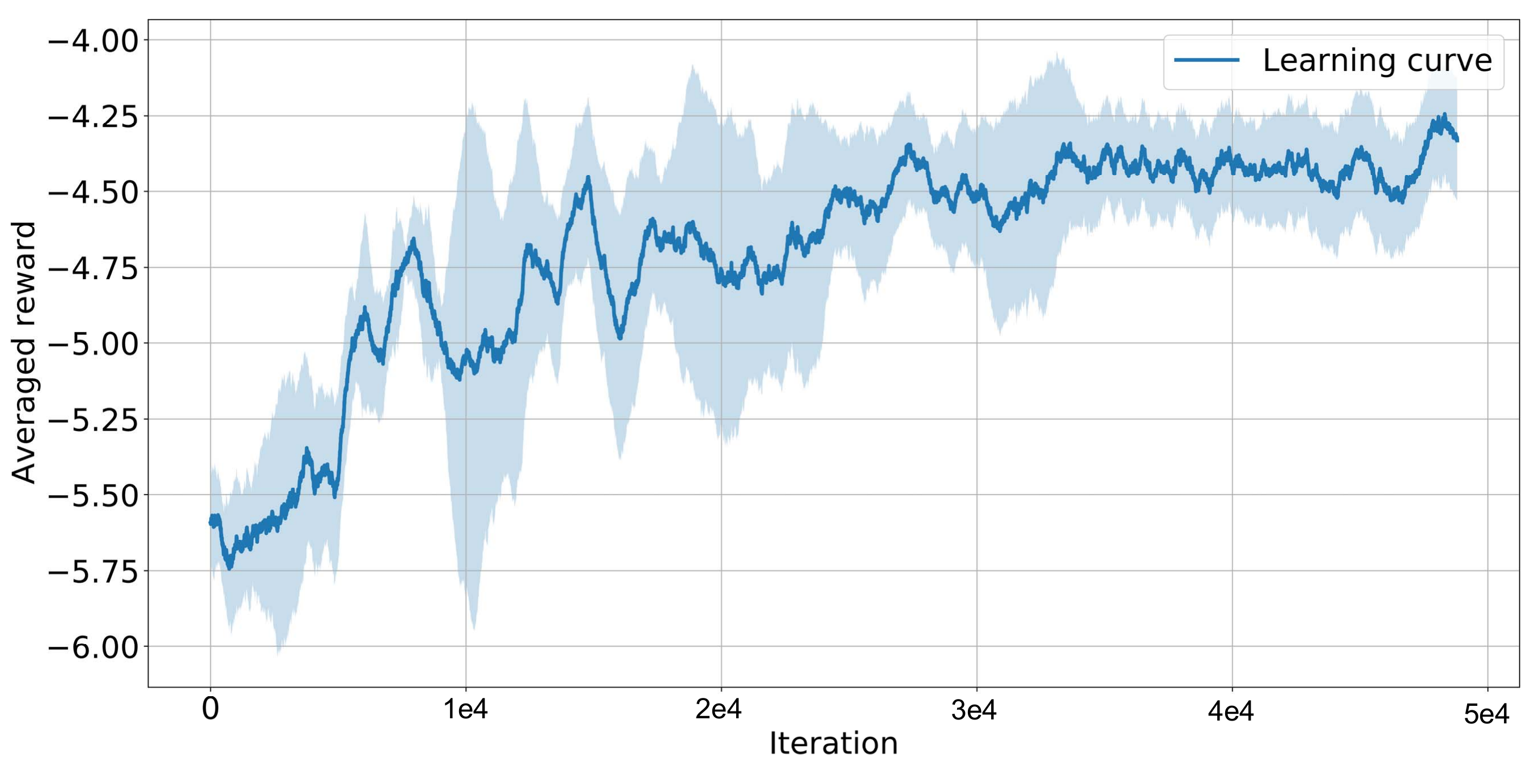}\\
    \caption{ {The learning curve of our model under train disturbances.}}
    \label{learning_curve_sd}
\end{center}
\end{figure}
\begin{table}[!b]
\renewcommand\arraystretch{1.2}
\caption{{The TTR results under train disturbances}}
\label{P_quality_sd}
\centering
\setlength{\tabcolsep}{2.1mm}{
    \begin{tabular}{c l c c c c}
        \toprule
        \multirow{2}{*}{Methods} &  & Test & \multicolumn{3}{c}{Generalization} \\ 
         & Metrics & \thead[c]{10*10}& \thead[c]{15*15} & \thead[c]{20*20} & \thead[c]{20*30} \\ 
        \midrule
         \multirow{3}{*}{FCFS} & Obj.(min)         & 1064           & 2295             & 3262             & 5756   \\ 
                               & Gap            & 68.42\%       & 76.30\%          & 80.41\%            & 88.67\%    \\ 
                               & Time(s)         & 0.086          &  0.165           &  0.270           & 0.395     \\
                               \midrule

         \multirow{3}{*}{FSFS} & Obj.(min)          & 657        & 1212         & 1361         & 1817   \\ 
                               & Gap          & 48.86\%        & 55.12\%          & 53.05\%          & 64.12\%   \\ 
                               & Time(s)         & \textbf{0.071} & \textbf{0.117}   & \textbf{0.133}   & \textbf{0.249}   \\
                               \midrule
\multirow{3}{*}{RLG}  & Obj.(min)  & 403     &  670             & 807                 & 842   \\ 
                               & Gap          & 16.63\%         &  18.81\%             &  20.82\%       & 22.57\%   \\ 
                               & Time(s)         & 0.45           &  1.598            & 2.044            & 3.379  \\ 
                               \midrule
          \multirow{3}{*}{RLG-LC}  & Obj.(min)  & 374     &  610             & 753              & 769   \\ 
                               & Gap             & 10.16\%       &  10.82\%        &  15.14\%            & 15.21\%  \\ 
                               & Time(s)         & 2.176          &  3.131           & 3.737             & 4.295  \\ 
                               \midrule
          \multirow{3}{*}{\makecell[c]{Gurobi\\(Baselines)}}  & Obj.(min)      
                              & \textbf{336}            & \textbf{544}              & \textbf{639}      & \textbf{652}    \\
                              & Opt.Rate         & 100\%          & 100\%            & 100\%            & 100\%   \\
                              & Time(s)          & 1.058          &  4.801           & 12.827           & 25.353   \\ 
        \bottomrule
\end{tabular}}
\end{table}
\subsection{TTR for Train Disturbances}
 {Given that the problem size (i.e., the number of decision variables) mainly depends on the number of stations ($I$) and trains ($K$), we denote the problem size as $I*K$. To verify the scalability of our approach, we first train our model on small-size instances denoted as 10*10, where 10 trains need to be dispatched at 10 stations. Subsequently, the learned model is allowed to dispatch the problem instances of a large scale. Its training process took about 6.25 hours, and Fig.~\ref{learning_curve_sd} shows the averaged learning curve over five independent training processes.} To test whether the RL model has learned dispatching skills, we compare the learned model with some handcrafted rules used in TTR, including First-Come-First-Service (FCFS) {~\cite{Wang2016}} and First-Schedule-First-Service (FSFS) {~\cite{Fang2015}}.  {Besides, we also compare our method with the exact algorithm implemented using Gurobi~\cite{Gurobi2022}. The performance of the learned model on 100 instances is shown in Table~\ref{P_quality_sd}.}

In some cases, the FSFS strategy may generate infeasible solutions due to the track capacity constraints, so we approximate its performance by using the remaining feasible solutions. 
The proposed RL-based method with graph representation (RLG) outperforms all handcrafted methods, which means that the proposed method is effective in learning skills via trial and error. Meanwhile, the learned model, similar to the hand-made rules, fits well with the real-time requirements of the TTR task. Its scalability can also be proved by testing larger-scale problems, including 15 stations and 15 trains (15*15), 20 stations and 20 trains (20*20), 20 stations and 30 trains (20*30). 
With the help of the local search, the proposed method (RLG-LC) can get a significant improvement in solution quality with small computational costs. Besides, we find that Gurobi, as one of the state-of-the-art solvers, also presents an excellent performance with an acceptable computation time when facing train disturbances.

\subsection{ {TTR for Train Disruptions}}   
\begin{figure}[!b]    
\begin{center}
    \includegraphics[width=0.48\textwidth]{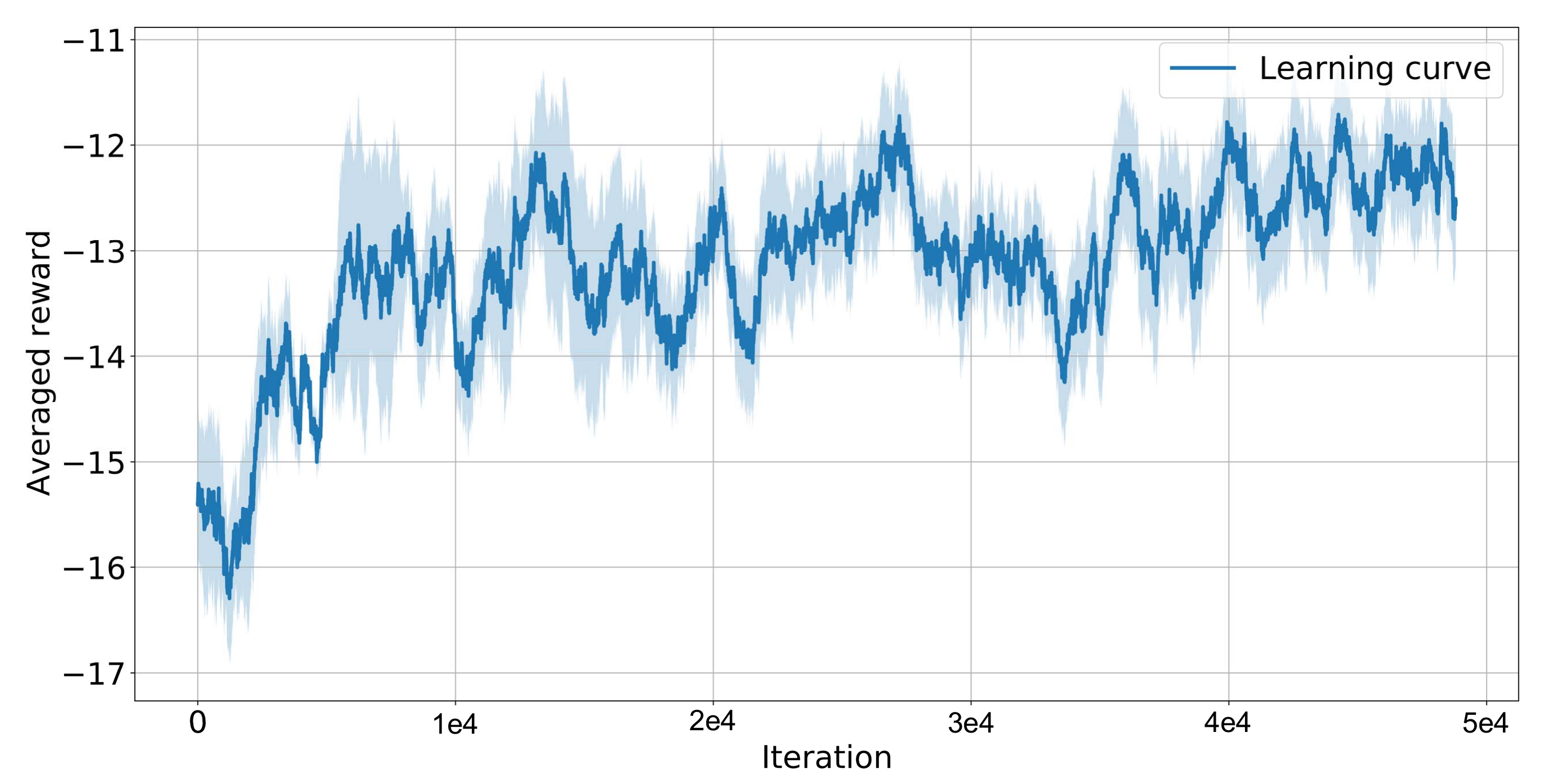}\\
    \caption{ {The learning curve of our model under train disruptions.}}
    \label{learning_curve2}
\end{center}
\end{figure}

 {With the help of the learning curriculum, our trained model can effectively address the TTR problem, even in cases of train disruptions. The associated learning curve is illustrated in Fig.~\ref{learning_curve2}.}  {Table~\ref{P_quality} shows the experimental results of the learned model (named RGL) in terms of solution quality and computation time on 100 instances.} 
To validate the effectiveness of the learning curriculum, we employ the vanilla PPO algorithm again to train our model in the complex scenarios, referred to as $\text{RLG}_\text{L}$, and also use the model learned in small-delay scenarios, denoted as $\text{RLG}_\text{S}$, directly in these complex scenarios. It is noteworthy that the performance of FSFS is not shown in this table, since it failed to find any feasible solution in almost all cases.

\begin{table}[htbp]
\renewcommand\arraystretch{1.2}
\caption{{The TTR results under train disruptions}}
\label{P_quality}
\centering
\setlength{\tabcolsep}{2.2mm}{
    \begin{tabular}{c l c c c c}
        \toprule
        \multirow{2}{*}{Methods} &  & Test & \multicolumn{3}{c}{Generalization} \\ 
         & \multirow{2}{*}{Metrics} & \thead[c]{10*10}& \thead[c]{15*15} & \thead[c]{20*20} & \thead[c]{20*30} \\ 
        \midrule
         \multirow{3}{*}{FCFS} & Obj.(min)        & 4792     & 10722      & 18407       & 24346   \\ 
                               & Gap              & 16.24\%  & 35.10\%    & 46.26\%     & 61.83\%    \\ 
                               & Time (s)   &\textbf{0.059}  &  \textbf{0.11}  & \textbf{0.179}   & \textbf{0.263}   \\
                                                    \midrule
        \multirow{3}{*}{$\text{RLG}_\text{S}$}    
                       & Obj. (min)       & 6073     & 10416       & 14449       & 12869   \\ 
                       & Gap              & 33.90\%   &  33.19\%  &  31.54\%    &  27.79\%  \\ 
                       & Time (s)         & 0.123     & 0.282      & 0.469       & 0.863 \\ 
                                            \midrule
        \multirow{3}{*}{$\text{RLG}_\text{L}$}    
                               & Obj. (min)       & 4711     & 10720       & 18780       & 24849   \\ 
                               & Gap              & 14.80\%   &  35.08\%  &  47.33\%     &  62.60\%   \\ 
                               & Time (s)         & 0.136     & 0.351      & 0.608       &1.174 \\ 
                                                    \midrule
         \multirow{3}{*}{RGL} 
                               & Obj. (min)       & 4260     & 7823       & 11758       & 12611   \\ 
                               & Gap              & 5.77\%   &  11.04\%   &  15.87\%  &  26.31\%   \\ 
                               & Time (s)         & 0.122     & 0.263      & 0.438  & 0.848 \\ 
                                                    \midrule
         \multirow{3}{*}{$\text{RGL\text{-}LC}$} 
                               & Obj. (min)     & 4187     & \textbf{7533}       & \textbf{11108}       & \textbf{11576}   \\
                               & Gap            &  4.13\%  &  7.62\%    &  10.95\%     & 19.72\%   \\ 
                               &Time (s)        & 2.53    &  4.188     & 5.245         & 5.897  \\ 
                                                    \midrule

          \multirow{3}{*}{Gurobi}   
                               & Obj. (min)     & \textbf{4015}     & 7629       & 15990        & 19874   \\
                               & Gap            & 1.27\%   &  8.78\%   &  38.14\%     &  53.24\%   \\ 
                               & Time (s)       & 1.27     &  5.937     & 8.264          & 10.214  \\ 
                        \midrule   
          \multirow{3}{*}{Baselines}  
                               & Obj. (min)     & 4014     & 6959       & 9892        & 9293    \\
                               & Opt.Rate       &  100 \%  &  100 \%    &  100\%      &  98\%   \\   
                               & Time (s)       & 2.03     &  58.63     & 714         & 1116  \\ 
        \bottomrule
\end{tabular}}
\end{table}

 Regarding solution quality, we can observe that the learned model $\text{RLG}_\text{S}$ does not perform as well as it did under train disturbances, and $\text{RLG}_\text{L}$ performs comparably to the FCFS strategy, especially on larger-scale problems. By contrast, $\text{RGL}$ achieves promising results, outperforming even the Gurobi solver in large-scale problems when using a similar computation time. With the help of the local search, its performance can be further improved, corresponding to $\text{RGL\text{-}LC}$. In terms of computational efficiency, these learned models can satisfy the real-time requirements of TTR in high-speed railways due to low computational costs. Even though the local search will increase the computation overhead to some extent, its response time is still acceptable for practical applications.
\begin{table}[!b]
\renewcommand\arraystretch{1.2}
\centering
\caption{ {Performance comparison of decision models with different state representation in terms of total arrival delay $J$ }}
\label{test_ablation}
\setlength{\tabcolsep}{4mm}{
    \begin{tabular}{c c c c c}
        \toprule
         Problem size & $\text{RLG}_{\text{raw}}$ & $\text{RLG}_{\text{1}}$ & $\text{RLG}_{\text{2}}$  & $\text{RLG}_{\text{AG}}$\\ 
        \midrule
        10*10 &  4991  & \textbf{4841}  & 5264  &  5073    \\ 
        15*15 &  9849 & \textbf{8143}   & 9675 &  9207   \\ 
        20*20 &  16852 & \textbf{12149} & 14831 &  13805   \\ 
        20*30 &  19540 & \textbf{13210} & 17463 &  14785   \\ 
        \bottomrule
\end{tabular}}
\end{table}

\begin{table*}[!t]
\renewcommand\arraystretch{1.2}
\centering
\caption{ {Performance comparison of decision models with different learning strategies in terms of total arrival delay $J$ }}
\label{ablation_rt}
\setlength{\tabcolsep}{3.3mm}{
    \begin{tabular}{c c c c c c c c c c}
        \toprule
         \multirow{2}{*}{Problem size} & \multirow{2}{*}{$\text{RLG}_\text{L}$} & \multirow{2}{*}{RGS} & \multicolumn{7}{c}{RGL}\\ 
         & & & $w_4$=0.1 &$w_4$=0.5  & $w_4$=1 &$w_4$=1.5 & $w_4$=2 & $w_4$=5 & $w_4$=10 \\ 
        \midrule
        10*10 &  4711 &  4352   &  4379  & \textbf{4260}& 4282   & 4282& 4301 & 4316 & 4427\\ 
        15*15 &  10720 &  7974   & 8114   & 7823 & \textbf{7805}  &7849& 7902 & 8126   & 8449\\ 
        20*20 &  18780 &  12093  &  12641 & \textbf{11758}& 11896 & 11988 & 12128 & 12492 & 12944\\ 
        20*30 &  24849 &  12997  &  15910 & \textbf{12611}& 12781 &12895  & 12829 & 12734 & 12812\\ 
        \bottomrule
\end{tabular}}
\end{table*}
 
\subsection{Ablation Studies}  
\subsubsection{\textbf{GNN-based state representation}}  	 
To verify the effectiveness of the graph representations for TTR, we trained different models using the following methods: (1) using the state consisting of raw features in the graph, denoted as $\text{RLG}_{\text{raw}}$; (2) employing the extracted state from different numbers of the GNN layers, denoted as $\text{RLG}_{n}$, where $n$ represents the number of GNN layers. (3) using an off-the-shelf graph (i.e., the alternative graph (AG)) {~\cite{alternative_graph}} to describe state, rather than the graph we designed, named $\text{RLG}_{\text{AG}}$.
 {We evaluated the performance of these models in terms of total arrival delays $J$ of solutions, calculated by Eq.~\eqref{objective},}  {and the experimental results on 100 instances are shown in Table~\ref{test_ablation}.}


It can be found that the performance of the decision model is improved with the help of GNN layers. In comparison to the model using raw features (\textit{i.e.}, $\text{RLG}_{\text{raw}}$), the model with one GNN layer (\textit{i.e.},$\text{RLG}_{\text{1}}$) consistently outperforms it on problems of all sizes. 
Although the performance of the decision model with two GNN layers (\textit{i.e.}, $\text{RLG}_{\text{2}}$) does not surpass that of $\text{RLG}_{\text{raw}}$ on the 10*10 problem, the influence of the GNN layer can still be observed on larger-scale problems. One possible reason is that when solving large-scale problems, the GNN can extract a more comprehensive perspective of the problem through the aggregation of node information. This global view of the problem may contribute to the decision model's ability to make better-informed decisions.
Additionally, by comparing $\text{RLG}_{\text{1}}$ with $\text{RLG}_{\text{AG}}$, the superiority of our designed graph for TTR can be demonstrated. This is because we modify the direction of edges when the two nodes are from the same trains, such that the edge points from the latter event to the former one. As a result, the decision model can access future information, such as predicted train delays at following stations, when determining the priority between two trains at the current station.
\subsubsection{\textbf{Learning Curriculum}}	
The proposed learning curriculum comprises two components, instance generation and loss definition. To assess their individual contributions, we initially trained our model using solely the sampling strategy, namely RGS. Subsequently, we examine the effect of the defined loss function by adjusting weight $w_4$ assigned to the loss item $L_4$.  {We present the experimental results on 100 instances in Table~\ref{ablation_rt}, } {where the performance is also defined as objective $J$.}

As indicated in the table, the $\text{RGS}$ model outperforms $\text{RLG}_\text{L}$, suggesting that our learning curriculum indeed enhances learning effectiveness. By selecting a value within the appropriate range of [0.5, 1.5] for hyperparameter $w_4$, the benefit of knowledge transfer can be found in the comparison between $\text{RGS}$ and $\text{RGL}$.
\subsubsection{\textbf{Local search}}		
\begin{figure}[!t]    
\begin{center}
    \includegraphics[width=0.48\textwidth]{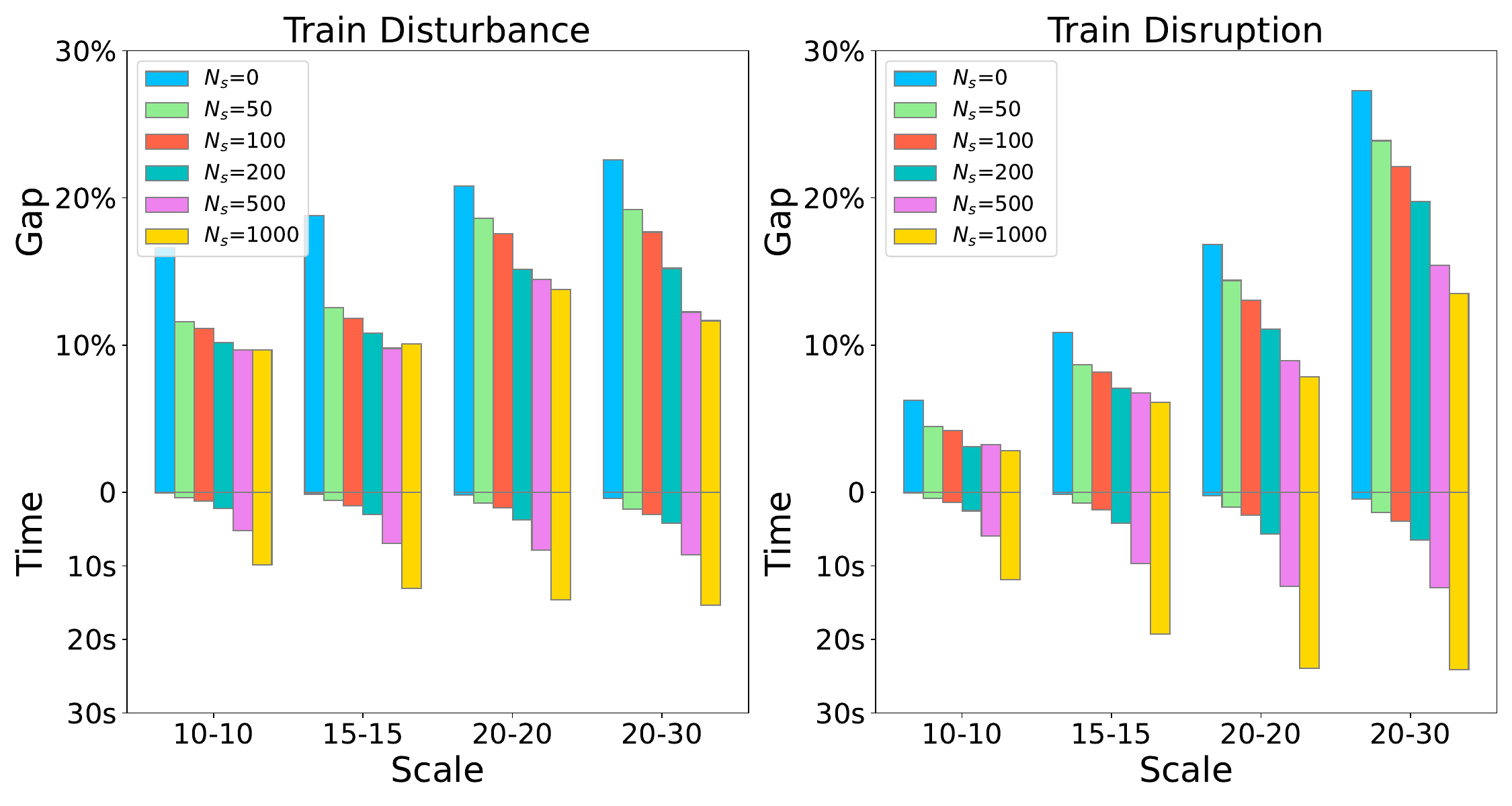}\\
    \caption{A comparison of performance in terms of local search times.}
    \label{dif_sample}
\end{center}
\end{figure}

In this part, we examine the influence of local search on our approach. Specifically, we vary the search times $N_s$ across different values, namely 0, 50, 100, 200, 500, and 1000.  {Fig.~\ref{dif_sample} shows the comparative results on problems with different scales, each evaluated on 100 instances.}

 {The experimental results underscore the beneficial influence of employing local search to enhance the model's scalability. Specifically, when the learned model is applied in isolation ($N_s=0$), its performance inevitably diminishes as the problem size increases, particularly in the case of train disruptions, but this issue can be alleviated to some degree by using local search. Especially, when the search times are set to 1000, the gap to the optimal solution can be reduced by up to 13\%. Besides, the computation cost of local search is also acceptable for practical application. Even with 1000 search iterations, its computation time is less than 30 seconds. Furthermore, the advantages of local search become more apparent with the increase of the problem size or the severity of train delays. It enhances our confidence in applying our approach across diverse delay scenarios.}

\section{Conclusion}
\label{Sec_conclusion}
In the daily operation of high-speed railways, providing a high-quality solution within a limited period is a grand challenge. In this study, we present a learning-based solution for TTR that directly satisfies the dispatcher's requirements for both response time and solution quality. In comparison to existing dispatching rules, the learned dispatching knowledge in our decision model can always obtain better performance across various problems with different scales and varying degrees of train delay, while maintaining reasonable computational costs. In particular, our method demonstrates superior performance over the Gurobi optimizer, a state-of-the-art solver, within a reasonable time frame when solving complex scenarios involving train disruptions, highlighting the practical value of the proposed approach in these challenging scenarios. Moreover, our ablation studies demonstrate the benefits of the GNN-based state representation, learning curriculum and local search, further underscoring the potential of our approach in tackling the TTR problem.

However, it should be noted that there is still a significant amount of work that remains to be done. For instance, in cases of train disruptions, dispatchers may resort to more aggressive tactics to recover train delays, such as canceling trains and breaking connections. Although these decisions are infrequent, they are crucial, and this study has not taken them into account. A potential solution is to develop a multi-agent learning system in which each agent is responsible for a specific type of action, which will be our future work. 

\section*{Acknowledgments}
This work is supported in part by the National Natural Science Foundation of China under Grants 61790574, 61773111, and U1834211, and in part by the Science and Technology Project of China National Railway Group Corporation Limited under Grant N2019G020. The work was done when YJ was with the Faculty of Technology, Bielefeld University, Germany funded by an Alexander von Humboldt Professorship for AI.

\bibliographystyle{IEEEtran}
\bibliography{scalable_TTR_3_12.bib}

\begin{thebibliography}{10}
\providecommand{\url}[1]{#1}
\csname url@samestyle\endcsname
\providecommand{\newblock}{\relax}
\providecommand{\bibinfo}[2]{#2}
\providecommand{\BIBentrySTDinterwordspacing}{\spaceskip=0pt\relax}
\providecommand{\BIBentryALTinterwordstretchfactor}{4}
\providecommand{\BIBentryALTinterwordspacing}{\spaceskip=\fontdimen2\font plus
\BIBentryALTinterwordstretchfactor\fontdimen3\font minus
  \fontdimen4\font\relax}
\providecommand{\BIBforeignlanguage}[2]{{%
\expandafter\ifx\csname l@#1\endcsname\relax
\typeout{** WARNING: IEEEtran.bst: No hyphenation pattern has been}%
\typeout{** loaded for the language `#1'. Using the pattern for}%
\typeout{** the default language instead.}%
\else
\language=\csname l@#1\endcsname
\fi
#2}}
\providecommand{\BIBdecl}{\relax}
\BIBdecl
\renewcommand{\BIBentryALTinterwordstretchfactor}{4}

\bibitem{Zhou2020}
P.~Zhou, L.~Chen, X.~Dai, B.~Li, and T.~Chai, ``Intelligent prediction of train
  delay changes and propagation using rvflns with improved transfer learning
  and ensemble learning,'' \emph{IEEE Transactions on Intelligent
  Transportation Systems}, vol.~22, no.~12, pp. 7432--7444, 2020.

\bibitem{cavone2020mpc}
G.~Cavone, T.~van~den Boom, L.~Blenkers, M.~Dotoli, C.~Seatzu, and
  B.~De~Schutter, ``An mpc-based rescheduling algorithm for disruptions and
  disturbances in large-scale railway networks,'' \emph{IEEE Transactions on
  Automation Science and Engineering}, vol.~19, no.~1, pp. 99--112, 2020.

\bibitem{Cacchiani2014}
V.~Cacchiani, D.~Huisman, M.~Kidd, L.~Kroon, P.~Toth, L.~Veelenturf, and
  J.~Wagenaar, ``An overview of recovery models and algorithms for real-time
  railway rescheduling,'' \emph{Transportation Research Part B:
  Methodological}, vol.~63, pp. 15--37, 2014.

\bibitem{Cheng2017}
R.~Cheng, Y.~Song, D.~Chen, and L.~Chen, ``{Intelligent Localization of a
  High-Speed Train Using LSSVM and the Online Sparse Optimization Approach},''
  \emph{IEEE Transactions on Intelligent Transportation Systems}, vol.~18,
  no.~8, pp. 2071--2084, Aug 2017.

\bibitem{Pellegrini2015}
P.~Pellegrini, G.~Marliere, R.~Pesenti, and J.~Rodriguez, ``{RECIFE-MILP: An
  Effective MILP-Based Heuristic for the Real-Time Railway Traffic Management
  Problem},'' \emph{IEEE Transactions on Intelligent Transportation Systems},
  vol.~16, no.~5, pp. 2609--2619, Oct 2015.

\bibitem{Dotoli2014}
M.~Dotoli, N.~Epicoco, M.~Falagario, B.~Turchiano, G.~Cavone, and
  A.~Convertini, ``{A Decision Support System for real-time rescheduling of
  railways},'' in \emph{2014 European Control Conference (ECC)}, Jun 2014, pp.
  696--701.

\bibitem{fang2015survey}
W.~Fang, S.~Yang, and X.~Yao, ``A survey on problem models and solution
  approaches to rescheduling in railway networks,'' \emph{IEEE Transactions on
  Intelligent Transportation Systems}, vol.~16, no.~6, pp. 2997--3016, 2015.

\bibitem{rodriguez2007constraint}
J.~Rodriguez, ``A constraint programming model for real-time train scheduling
  at junctions,'' \emph{Transportation Research Part B: Methodological},
  vol.~41, no.~2, pp. 231--245, 2007.

\bibitem{Sato2013}
K.~Sato, K.~Tamura, and N.~Tomii, ``{A MIP-based timetable rescheduling
  formulation and algorithm minimizing further inconvenience to passengers},''
  \emph{Journal of Rail Transport Planning \& Management}, vol.~3, no.~3, pp.
  38--53, Aug 2013.

\bibitem{kersbergen2016distributed}
B.~Kersbergen, T.~van~den Boom, and B.~De~Schutter, ``Distributed model
  predictive control for railway traffic management,'' \emph{Transportation
  Research Part C: Emerging Technologies}, vol.~68, pp. 462--489, 2016.

\bibitem{Zhang2020a}
C.~Zhang, Y.~Gao, L.~Yang, Z.~Gao, and J.~Qi, ``{Joint optimization of train
  scheduling and maintenance planning in a railway network: A heuristic
  algorithm using Lagrangian relaxation},'' \emph{Transportation Research Part
  B: Methodological}, vol. 134, pp. 64--92, Apr 2020.

\bibitem{Min2011}
Y.-H. Min, M.-J. Park, S.-P. Hong, and S.-H. Hong, ``An appraisal of a
  column-generation-based algorithm for centralized train-conflict resolution
  on a metropolitan railway network,'' \emph{Transportation Research Part B:
  Methodological}, vol.~45, no.~2, pp. 409--429, 2011.

\bibitem{Zhan2021}
S.~Zhan, S.~Wong, P.~Shang, Q.~Peng, J.~Xie, and S.~Lo, ``{Integrated railway
  timetable rescheduling and dynamic passenger routing during a complete
  blockage},'' \emph{Transportation Research Part B: Methodological}, vol. 143,
  pp. 86--123, Jan 2021.

\bibitem{Bettinelli2017}
A.~Bettinelli, A.~Santini, and D.~Vigo, ``{A real-time conflict solution
  algorithm for the train rescheduling problem},'' \emph{Transportation
  Research Part B: Methodological}, vol. 106, pp. 237--265, Dec 2017.

\bibitem{TornquistKrasemann2012}
J.~{T{\"{o}}rnquist Krasemann}, ``{Design of an effective algorithm for fast
  response to the re-scheduling of railway traffic during disturbances},''
  \emph{Transportation Research Part C: Emerging Technologies}, vol.~20, no.~1,
  pp. 62--78, Feb 2012.

\bibitem{dundar2013train}
S.~D{\"u}ndar and {\.I}.~{\c{S}}ahin, ``Train re-scheduling with genetic
  algorithms and artificial neural networks for single-track railways,''
  \emph{Transportation Research Part C: Emerging Technologies}, vol.~27, pp.
  1--15, 2013.

\bibitem{kanai2011optimal}
S.~Kanai, K.~Shiina, S.~Harada, and N.~Tomii, ``An optimal delay management
  algorithm from passengers’ viewpoints considering the whole railway
  network,'' \emph{Journal of Rail Transport Planning \& Management}, vol.~1,
  no.~1, pp. 25--37, 2011.

\bibitem{wang2019genetic}
M.~Wang, L.~Wang, X.~Xu, Y.~Qin, and L.~Qin, ``Genetic algorithm-based particle
  swarm optimization approach to reschedule high-speed railway timetables: a
  case study in china,'' \emph{Journal of Advanced Transportation}, vol. 2019,
  2019.

\bibitem{Wang2016}
P.~Wang, L.~Ma, R.~M.~P. Goverde, and Q.~Wang, ``{Rescheduling Trains Using
  Petri Nets and Heuristic Search},'' \emph{IEEE Transactions on Intelligent
  Transportation Systems}, vol.~17, no.~3, pp. 726--735, Mar 2016.

\bibitem{Fang2015}
W.~Fang, S.~Yang, and X.~Yao, ``{A Survey on Problem Models and Solution
  Approaches to Rescheduling in Railway Networks},'' \emph{IEEE Transactions on
  Intelligent Transportation Systems}, vol.~16, no.~6, pp. 2997--3016, Dec
  2015.

\bibitem{Semrov2016}
D.~{\v{S}}emrov, R.~Marseti{\v{c}}, M.~{\v{Z}}ura, L.~Todorovski, and A.~Srdic,
  ``{Reinforcement learning approach for train rescheduling on a single-track
  railway},'' \emph{Transportation Research Part B: Methodological}, vol.~86,
  pp. 250--267, Apr 2016.

\bibitem{Li2022}
W.~Li and S.~Ni, ``{Train timetabling with the general learning environment and
  multi-agent deep reinforcement learning},'' \emph{Transportation Research
  Part B: Methodological}, vol. 157, pp. 230--251, Mar 2022.

\bibitem{Khadilkar2019}
H.~Khadilkar, ``A scalable reinforcement learning algorithm for scheduling
  railway lines,'' \emph{IEEE Transactions on Intelligent Transportation
  Systems}, vol.~20, no.~2, pp. 727--736, 2018.

\bibitem{Zhu2020}
Y.~Zhu, H.~Wang, and R.~M. Goverde, ``{Reinforcement Learning in Railway
  Timetable Rescheduling},'' in \emph{2020 IEEE 23rd International Conference
  on Intelligent Transportation Systems (ITSC)}, 2020, pp. 1--6.

\bibitem{9564980}
Y.~Wang, Y.~Lv, J.~Zhou, Z.~Yuan, Q.~Zhang, and M.~Zhou, ``A policy-based
  reinforcement learning approach for high-speed railway timetable
  rescheduling,'' in \emph{2021 IEEE International Intelligent Transportation
  Systems Conference (ITSC)}, 2021, pp. 2362--2367.

\bibitem{Ning2021}
L.~Ning, Y.~Li, M.~Zhou, H.~Song, and H.~Dong, ``A deep reinforcement learning
  approach to high-speed train timetable rescheduling under disturbances,'' in
  \emph{2019 IEEE Intelligent Transportation Systems Conference (ITSC)}, 2019,
  pp. 3469--3474.

\bibitem{yue2023reinforcement}
P.~Yue, Y.~Jin, X.~Dai, Z.~Feng, and D.~Cui, ``Reinforcement learning for
  online dispatching policy in real-time train timetable rescheduling,''
  \emph{IEEE Transactions on Intelligent Transportation Systems}, 2023.

\bibitem{kool2018attention}
W.~Kool, H.~Van~Hoof, and M.~Welling, ``Attention, learn to solve routing
  problems!'' \emph{arXiv preprint arXiv:1803.08475}, 2018.

\bibitem{Bello2016}
I.~Bello, H.~Pham, Q.~V. Le, M.~Norouzi, and S.~Bengio, ``{Neural Combinatorial
  Optimization with Reinforcement Learning},'' \emph{5th International
  Conference on Learning Representations, ICLR 2017 - Workshop Track
  Proceedings}, pp. 1--15, nov 2016.

\bibitem{hottung2021}
A.~Hottung, Y.-D. Kwon, and K.~Tierney, ``Efficient active search for
  combinatorial optimization problems,'' \emph{arXiv preprint
  arXiv:2106.05126}, 2021.

\bibitem{Zhang2020c}
K.~Zhang, F.~He, Z.~Zhang, X.~Lin, and M.~Li, ``Multi-vehicle routing problems
  with soft time windows: A multi-agent reinforcement learning approach,''
  \emph{Transportation Research Part C: Emerging Technologies}, vol. 121, p.
  102861, 2020.

\bibitem{zhu2017bi}
Y.~Zhu, B.~Mao, Y.~Bai, and S.~Chen, ``A bi-level model for single-line rail
  timetable design with consideration of demand and capacity,''
  \emph{Transportation Research Part C: Emerging Technologies}, vol.~85, pp.
  211--233, 2017.

\bibitem{Wang2021}
Y.~Wang, Y.~Lv, J.~Zhou, Z.~Yuan, Q.~Zhang, and M.~Zhou, ``A policy-based
  reinforcement learning approach for high-speed railway timetable
  rescheduling,'' in \emph{2021 IEEE International Intelligent Transportation
  Systems Conference (ITSC)}, 2021, pp. 2362--2367.

\bibitem{xu2018powerful}
K.~Xu, W.~Hu, J.~Leskovec, and S.~Jegelka, ``How powerful are graph neural
  networks?'' \emph{arXiv preprint arXiv:1810.00826}, 2018.

\bibitem{Zhang2020b}
C.~Zhang, W.~Song, Z.~Cao, J.~Zhang, P.~S. Tan, and X.~Chi, ``Learning to
  dispatch for job shop scheduling via deep reinforcement learning,'' in
  \emph{Advances in Neural Information Processing Systems}, vol.~33, 2020, pp.
  1621--1632.

\bibitem{wang2018passenger}
Y.~Wang, A.~D’Ariano, J.~Yin, L.~Meng, T.~Tang, and B.~Ning, ``Passenger
  demand oriented train scheduling and rolling stock circulation planning for
  an urban rail transit line,'' \emph{Transportation Research Part B:
  Methodological}, vol. 118, pp. 193--227, 2018.

\bibitem{costa20a}
P.~R. d.~O. da~Costa, J.~Rhuggenaath, Y.~Zhang, and A.~Akcay, ``Learning 2-opt
  heuristics for the traveling salesman problem via deep reinforcement
  learning,'' in \emph{Proceedings of The 12th Asian Conference on Machine
  Learning}, vol. 129, Nov 2020, pp. 465--480.

\bibitem{Gurobi2022}
``Gurobi optimizer reference manual,'' \url{http://www. gurobi. com.}, 2022.

\bibitem{alternative_graph}
F.~Corman, A.~D'Ariano, D.~Pacciarelli, and M.~Pranzo, ``{A tabu search
  algorithm for rerouting trains during rail operations},''
  \emph{Transportation Research Part B: Methodological}, vol.~44, no.~1, pp.
  175--192, Jan 2010.

\end{thebibliography}

\vfill
\end{document}